\documentclass{article}

\usepackage{lmodern}
\usepackage{PRIMEarxiv}
\usepackage{arydshln}
\usepackage{xcolor}
\usepackage{comment}
\usepackage{arydshln}
\usepackage[utf8]{inputenc} 
\usepackage[T1]{fontenc}    
\usepackage{hyperref}       
\usepackage{url}
\usepackage{pifont}  
\usepackage{booktabs}       
\usepackage{amsfonts}       
\usepackage{nicefrac}       
\usepackage{microtype}      
\usepackage{lipsum}
\usepackage{fancyhdr}       
\usepackage{graphicx}       
\graphicspath{{media/}}     
\usepackage{float}
\usepackage{subcaption} 
\usepackage{adjustbox}
\usepackage{caption}
\usepackage{natbib} 
\usepackage{graphicx} 
\usepackage{amsmath}

\title{Supervised Contrastive Learning for Few-Shot AI-Generated Image Detection and Attribution
\thanks{\textit{\underline{Citation}}: 
\textbf{Authors. Title. Pages.... DOI:000000/11111.}} 
}

\author{
  Jaime Álvarez Urueña,  \\
  Universidad Politécnica de Madrid (UPM) \\
  Madrid\\
  \texttt{jaime.alvarez.uruena@alumnos.upm.es} \\
   \And
  Javier Huertas Tato, David Camacho \\
  Universidad Politécnica de Madrid (UPM) \\
  Madrid\\
\texttt{\{javier.huertas.tato, david.camacho\}@upm.es} \\
}

\begin{document}
\maketitle

\begin{abstract}
The rapid advancement of generative artificial intelligence has enabled the creation of synthetic images that are increasingly indistinguishable from authentic content, posing significant challenges for digital media integrity. This problem is compounded by the accelerated release cycle of novel generative models, which renders traditional detection approaches (reliant on periodic retraining) computationally infeasible and operationally impractical. The proliferation of new generative architectures creates a fundamental scalability challenge wherein detection systems rapidly become obsolete.

This work proposes a novel two-stage detection framework designed to address the generalization challenge inherent in synthetic image detection. The first stage employs a vision deep learning model trained via supervised contrastive learning to extract discriminative embeddings from input imagery. Critically, this model was trained on a strategically partitioned subset of available generators, with specific architectures withheld from training to rigorously ablate cross-generator generalization capabilities. The second stage utilizes a k-nearest neighbors (k-NN) classifier operating on the learned embedding space, trained in a few-shot learning paradigm incorporating limited samples from previously unseen test generators. This hybrid architecture combines deep metric learning with non-parametric classification to enhance adaptability to novel generative models.

Experimental results demonstrate that this approach substantially outperforms current state-of-the-art methods in synthetic image detection. With merely 150 images per class in the few-shot learning regime, which are easily obtainable from current generation models, the proposed framework achieves an average detection accuracy of 91.3\%, representing a 5.2 percentage point improvement over existing approaches while exhibiting superior stability and generalization to generators unseen during training. For the source attribution task, the proposed approach obtains improvements of  of 14.70\% and 4.27\% in AUC and OSCR respecively on an open set classification context, marking a significant advancement toward robust, scalable forensic attribution systems capable of adapting to the evolving generative AI landscape without requiring exhaustive retraining protocols.
\end{abstract}


\section{Introduction}
 Despite the immense benefits and transformative potential offered by generative AI, it also presents significant challenges and risks ~\cite{cite3}. For instance, AI-powered chatbots can be misused by students to complete academic assignments dishonestly \cite{cite7}. Additionally, generative image models have the potential to create misleading or controversial images that could influence political decisions \cite{cite8}. Similarly, AI-generated audio models pose risks related to copyright infringement and unauthorized content creation \cite{cite62}.

This work focuses on the detection of AI-generated images, which represents a significant threat as one of the most potentially harmful forms of fake news \cite{cite9}. While text generation can lead to issues such as copyright infringement or misinformation, the creation of AI-generated images is particularly concerning due to its potential detrimental effects. This worsens when images depict well-known individuals in misleading contexts, suggesting they have engaged in certain actions that may not be true \cite{peng2025crafting}.

\begin{figure}[htbp]

    \setlength{\tabcolsep}{1pt} 
    \renewcommand{\arraystretch}{1} 
    \begin{tabular}{ccccccccc}

    \raisebox{0.037\textwidth}{\rotatebox{90}{Real}} &
        \includegraphics[width=0.115\textwidth]{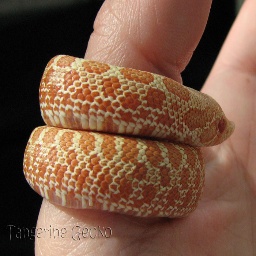}\hspace{-11pt} &
        \includegraphics[width=0.115\textwidth]{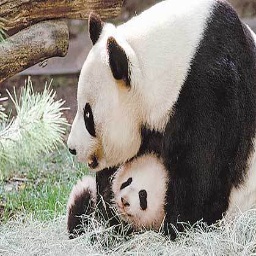} &
        \includegraphics[width=0.115\textwidth]{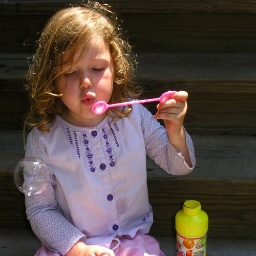} &
        \includegraphics[width=0.115\textwidth]{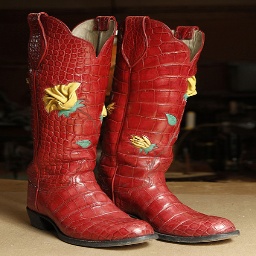} &
        \includegraphics[width=0.115\textwidth]{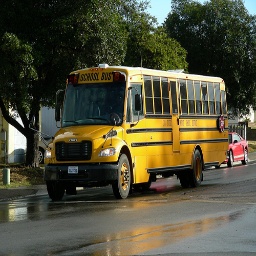} &
        \includegraphics[width=0.115\textwidth]{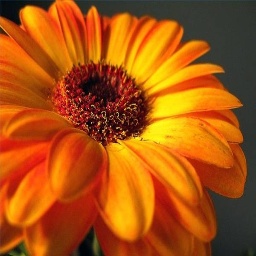} &
        \includegraphics[width=0.115\textwidth]{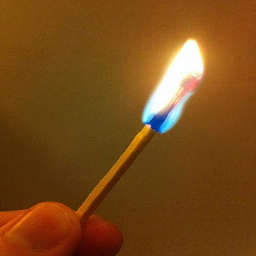} &
        \includegraphics[width=0.115\textwidth]{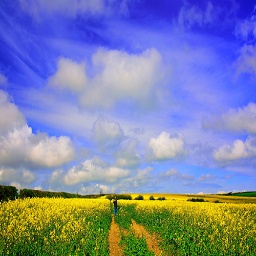}\\[-1pt]

       \raisebox{0.025\textwidth}{\rotatebox{90}{Synthetic}} &
        \includegraphics[width=0.115\textwidth]{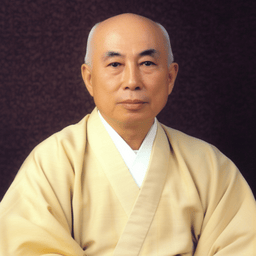}\hspace{-11pt} &
        \includegraphics[width=0.115\textwidth]{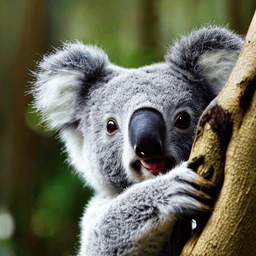} &
        \includegraphics[width=0.115\textwidth]{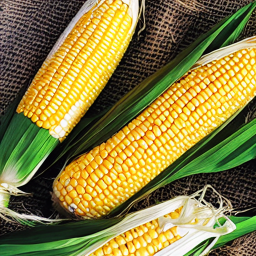} &
        \includegraphics[width=0.115\textwidth]{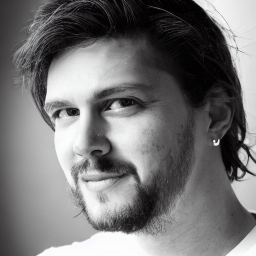} &
        \includegraphics[width=0.115\textwidth]{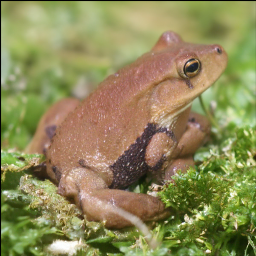} &
        \includegraphics[width=0.115\textwidth]{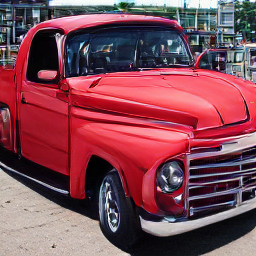} &
        \includegraphics[width=0.115\textwidth]{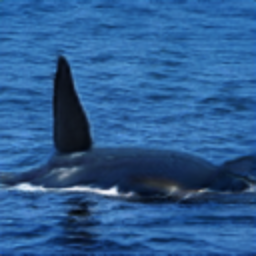} &
        \includegraphics[width=0.115\textwidth]{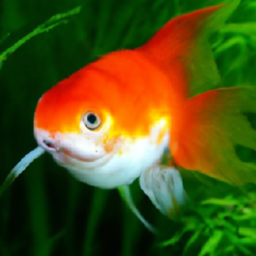} \\ 
        
        &Midjourney \cite{midjourney}&Wukong \cite{wukong}&SD\_1.4 \cite{stable_diffusion_1}&SD\_1.5 \cite{stable_diffusion_1}&ADM \cite{adm}&VQDM \cite{vqdm}&BigGan \cite{bigGan}&Glide \cite{glide}
    \end{tabular}
    \caption{Comparison of real images (top row) and synthetic images (bottom row). Each column shows images from a different synthetic image generator. Code and models developed for this work are available at \url{https://github.com/JaimeAlvarez18/SupConLoss_fake_image_detection}}
    \label{figure:1}
\end{figure}

Detecting AI-generated images is a complex and increasingly challenging task. In earlier stages of development, image generation models often produced outputs with evident artifacts and inconsistencies which made distinguishing synthetic images from real ones relatively straightforward \cite{zheng2024breaking,wang2020cnn}. However, the rapid advancement of generative models has significantly narrowed the perceptual gap between real and AI-generated content. Modern image generators can now produce realistic images across a broad range of domains, highlighting their versatility \cite{aziz2025global}. Figure \ref{figure:1} depicts real and fake images created by different generators. Without prior knowledge, even for a human being, it would be challenging to tell whether those images are synthetic and even more challenging to tell their original source.

The fast-paced development of new image generation models introduces another layer of complexity. The number of models capable of producing highly realistic images is growing dramatically \cite{cite4}, and, as a result, a model trained on current image generators may not remain effective in the near future, as new and more sophisticated image-generation techniques will likely emerge \cite{cite10}. In practical terms, it is not feasible to retrain detection systems each time a new generator is released. Consequently, it is critical for detection models to generalize effectively to generators that were not encountered during the training phase \cite{park2025community}. This requirement is particularly pressing given that many state-of-the-art generative models are proprietary or closed-source, making it difficult to obtain a sufficient number of training examples (typically exceeding 1,000) when using standard deep learning approaches. The limited availability of images per generator further constrains the training process, hindering the model’s ability to learn robust and generalizable features. Overcoming these challenges requires the development of detection methods that are both data-efficient and resilient to the rapid evolution of generative technologies. \cite{wufew}

In this work, a comprehensive research is conducted to evaluate various approaches for effectively distinguishing AI-generated images from real ones, as well as attributing each synthetic image to its generator. The proposed work in this paper makes a threefold contribution:

\begin{itemize}

\item A novel framework is introduced for training classifiers capable of distinguishing real from synthetic images and addressing the image source attribution problem. The proposed approach employs Supervised Contrastive Loss (SupConLoss) in conjunction with a MambaVision model to extract high-quality image embeddings. These embeddings are subsequently classified using a k-Nearest Neighbors (k-NN) algorithm within a few-shot learning paradigm, effectively leveraging a limited number of generator instances for training.

\item A comprehensive experimental analysis is conducted to evaluate the impact of varying the number of instances used in the few-shot learning configuration.

\item An explainability study is presented using LIME models to identify salient pixels that influence embedding assignments, complemented by clustered visualizations of the learned embedding space.

\end{itemize}

To present these contributions clearly, the remainder of this article is organized as follows: Section \ref{sota} reviews prior work, highlighting the challenges and solutions proposed by other authors. Section \ref{methods} describes the architectures, datasets, and experimental setup employed. Section \ref{results} presents the experimental results along with discussions on explainability and visualization results. Finally, Section \ref{conclusions} summarizes the conclusions and outlines directions for future work.

\section{Related Work}
\label{sota}
As image generators have advanced significantly in recent years, they are now capable of producing highly realistic images (often indistinguishable to the human eye) posing a potential threat in terms of misinformation, digital forgery, and content authenticity. Consequently, researchers have begun developing various approaches and techniques to detect synthetic images from real ones. Furthermore, this task can be extended beyond mere detection to address the source attribution problem, which consists on identifying which generator produced a given image. Such capabilities are valuable not only
for detecting emerging, previously unknown generators, but also for identifying when existing
detection models become outdated and require retraining with images from newer generation
techniques.

\subsection{Supervised methods for Synthetic Image Classification}
As these tasks involve image classification, early studies predominantly adopted supervised learning approaches \cite{cite13,cite19,cite30}. In \cite{cite13}, the authors conducted a series of experiments using the GenImage \cite{cite14} and DiffusionForensics \cite{cite15} datasets to evaluate their proposed architecture. Their experimental setup involved training the model exclusively on images generated by a single generator (Stable Diffusion v1.4) and subsequently testing it on images produced by multiple other generators. The proposed method focused on extracting artifact-based features from images to perform binary classification (real vs. fake). Artifact features refer to visual or statistical irregularities unintentionally introduced during the image synthesis process, essentially generation errors that serve as distinguishing cues. This artifact-based feature extraction strategy has been widely adopted in subsequent studies, as image generators, particularly Generative Adversarial Networks (GANs), tend to introduce a substantial number of such anomalies during the generation process \cite{cite20,cite21}.

Similarly, \cite{cite19} employed a traditional binary classification framework to differentiate between real and fake images. In addition, they utilized Gradient-weighted Class Activation Mapping (Grad-CAM) to analyze the model’s attention patterns. Their results indicated that for real images, the classifier leveraged broader and more semantically coherent regions, whereas for fake images, the model’s attention was fragmented and sparse, suggesting reliance on low-level inconsistencies.

The study presented in \cite{cite30} explored the generalization capability of detection models by training on images generated by N generators and testing on (N+k) unseen ones, following the chronological release order of image generation models. Their findings revealed that while fake image detection models exhibit a certain degree of cross-generator generalization, their performance significantly deteriorates when confronted with images from generators whose architectures differ substantially from those seen during training.

Despite these advancements, the primary limitation of supervised approaches lies in their poor generalization to unseen generators. This is a critical issue, as new image generators are being developed and released at an increasingly rapid pace. Consequently, retraining or fine-tuning a supervised classifier each time a new generator appears is impractical and unsustainable \cite{cite10}. In order to enhance the effectiveness and generalization of supervised approaches, self-supervised pretraining or contrastive representation learning could be integrated prior to supervised fine-tuning, allowing models to learn rich and transferable visual features without reliance on labeled fake data.

\subsection{Representation Learning Methods in Digital Image Forensics}

The subsequent line of research explored contrastive learning–based methods, encompassing both zero-shot and few-shot learning paradigms \cite{cite28,cite29,cite31,cite32}. Models trained under this paradigm demonstrate the potential to identify images generated by previously unseen models \cite{cite33}, thereby mitigating the dependence on large, diverse datasets of fake images from multiple generators.

In \cite{cite28}, the authors leveraged CLIP \cite{clip} to extract high-level semantic representations from images. To enhance the model’s sensitivity to generation artifacts, they applied patch-wise feature extraction, capturing local inconsistencies at the patch level that facilitate the discrimination between real and synthetic images. Similarly, \cite{cite29} utilized CLIP as an image encoder to generate embeddings for both real and fake images, which were subsequently classified using a Support Vector Machine (SVM).

In a different direction, \cite{cite30} proposed a zero-shot entropy-based detector (ZED) that required no fake images during training. Their approach computed statistical features from input images and compared them against distributions derived from real images to detect potential forgeries.

The study in \cite{cite31} introduced a LoRA-based Forgery Awareness Module, achieving a 14.55\% improvement in classification accuracy while utilizing only 0.56\% of the training data required by competing methods. This demonstrated the efficiency and adaptability of parameter-efficient fine-tuning techniques for forgery detection.

Finally, \cite{cite32} proposed a novel contrastive learning strategy that integrates Denoising Diffusion Implicit Models (DDIM) for image perturbation and reconstruction prior to classification. This approach was motivated by the empirical observation that synthetic images exhibit greater robustness than real ones when subjected to DDIM-based perturbations and reconstructions \cite{cite34}, providing a new avenue for distinguishing generated content from authentic imagery.

\subsection{Challenges in Cross-Generator generalization}

Among the various feature extraction techniques explored in prior works, forensic-based methods stand out for their interpretability and their ability to capture intrinsic patterns of different image generators. These methods typically involve computing specific statistical or frequency-domain metrics from authentic images and comparing them against those extracted from the images to be classified \cite{xu2025detecting,zhang2019detecting}.

In \cite{xu2025detecting}, the authors introduced an approach based on high-frequency perturbations applied to mid-level embeddings, enabling the detection not only of the source generator responsible for image synthesis, but also of the hyperparameters used during the process. Conversely, \cite{zhang2019detecting} employed the 2D Discrete Fourier Transform (DFT) on each of the RGB channels to obtain three corresponding frequency spectra. The logarithmic spectrum of each channel was then computed, normalized to the interval $[-1,1]$, and subsequently fed into a classification model.

While these forensic approaches provide rich information for distinguishing real and synthetic content, they exhibit limited generalization capabilities when faced with images produced by previously unseen generators. This limitation arises because the statistical characteristics of images vary significantly across different generation architectures \cite{xuan2019generalization,corvi2023intriguing}. As highlighted by \cite{tariang2024synthetic}, the fingerprints of generative models depend not only on the model architecture itself but also on the training dataset and hyperparameter configurations. Consequently, a detection model may accurately identify images from a generator trained under one setting but fail to recognize images from the same generator trained with different data or parameters. This underscores the substantial challenge of achieving cross-generator generalization in forensic-based detection.

To address this limitation, the approach proposed in this work adopts the opposite paradigm: rather than relying on predefined, fixed statistical metrics, the model is trained to learn discriminative features directly from the input images. This allows the network to autonomously discover feature representations that encode informative cues about the source generator, enabling it to recognize and cluster embeddings from previously unseen generation models \cite{zhao2024review}. This approach is based on the fact that deep learning does not need explicit features extracted from input data, as it itself is able to extract ad hoc features that are useful for the addressed problem \cite{cruciani2020feature}.

\section{Proposed Framework}
\label{methods}

\subsection{Motivation and Conceptual Overview}
Contrastive learning offers a key advantage over traditional supervised learning: its ability to classify instances of previously unseen classes during inference \cite{contr}. This property stems from contrastive learning specific focus on representations instead of labels, which enables the model to detect fake images from unknown sources. In contrast, supervised learning typically fails to recognize classes not encountered during training, as its final classification layer contains a fixed number of neurons corresponding to the predefined classes. Consequently, the model architecture must be modified (and possible retrained) before it can accommodate additional classes.

Moreover, contrastive learning has demonstrated superior performance across a wide range of tasks and datasets \cite{adv}. Unlike supervised learning, which optimizes for class membership, contrastive learning focuses on learning representations by bringing similar samples closer in the latent space and pushing dissimilar ones apart. As a result, contrastive models do not rely on ad hoc discriminative features for each class but instead learn structured, semantically meaningful embeddings that naturally cluster instances by similarity in the latent space.

\begin{figure}[htbp]
    \centering
    \begin{subfigure}[b]{0.32\textwidth}
        \centering
        \includegraphics[width=\linewidth]{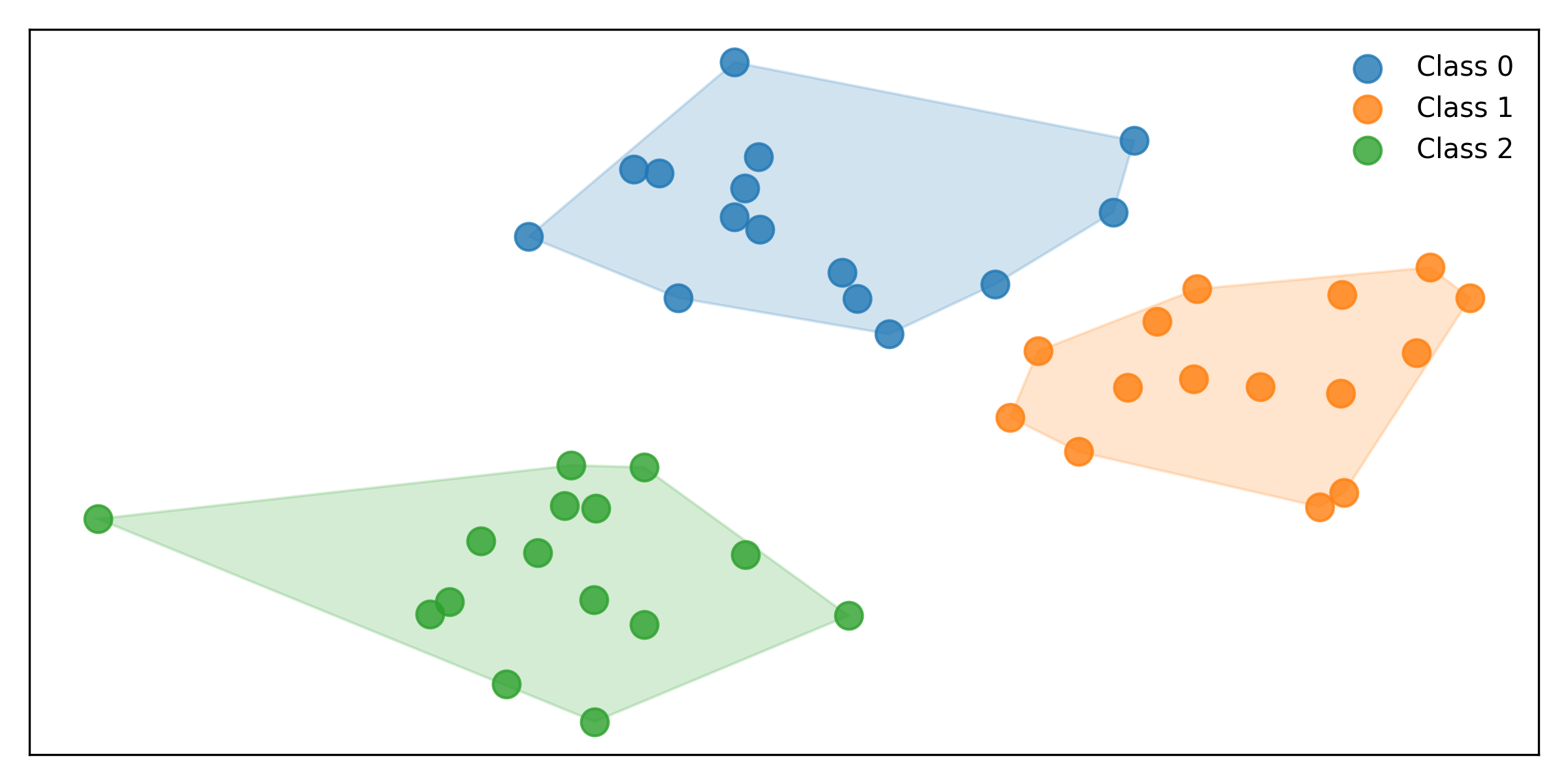}
        \caption{Supervised Learning (Decision boundaries).}
        \label{fig:sub1}
    \end{subfigure}
    \hfill
    \begin{subfigure}[b]{0.32\textwidth}
        \centering
        \includegraphics[width=\linewidth]{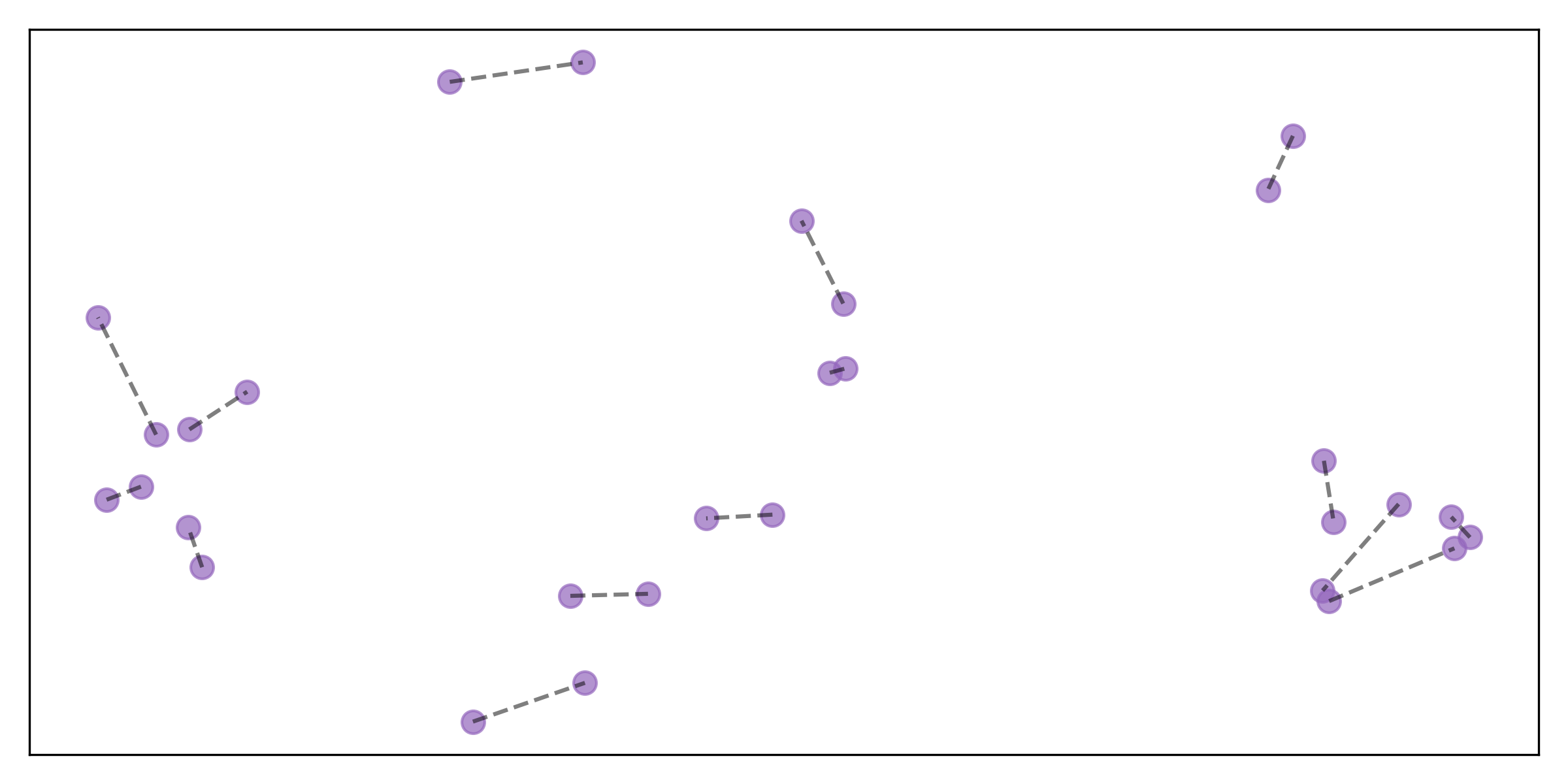}
        \caption{Contrastive Learning (Instance-level similarity).}
        \label{fig:sub2}
    \end{subfigure}
    \hfill
    \begin{subfigure}[b]{0.32\textwidth}
        \centering
        \includegraphics[width=\linewidth]{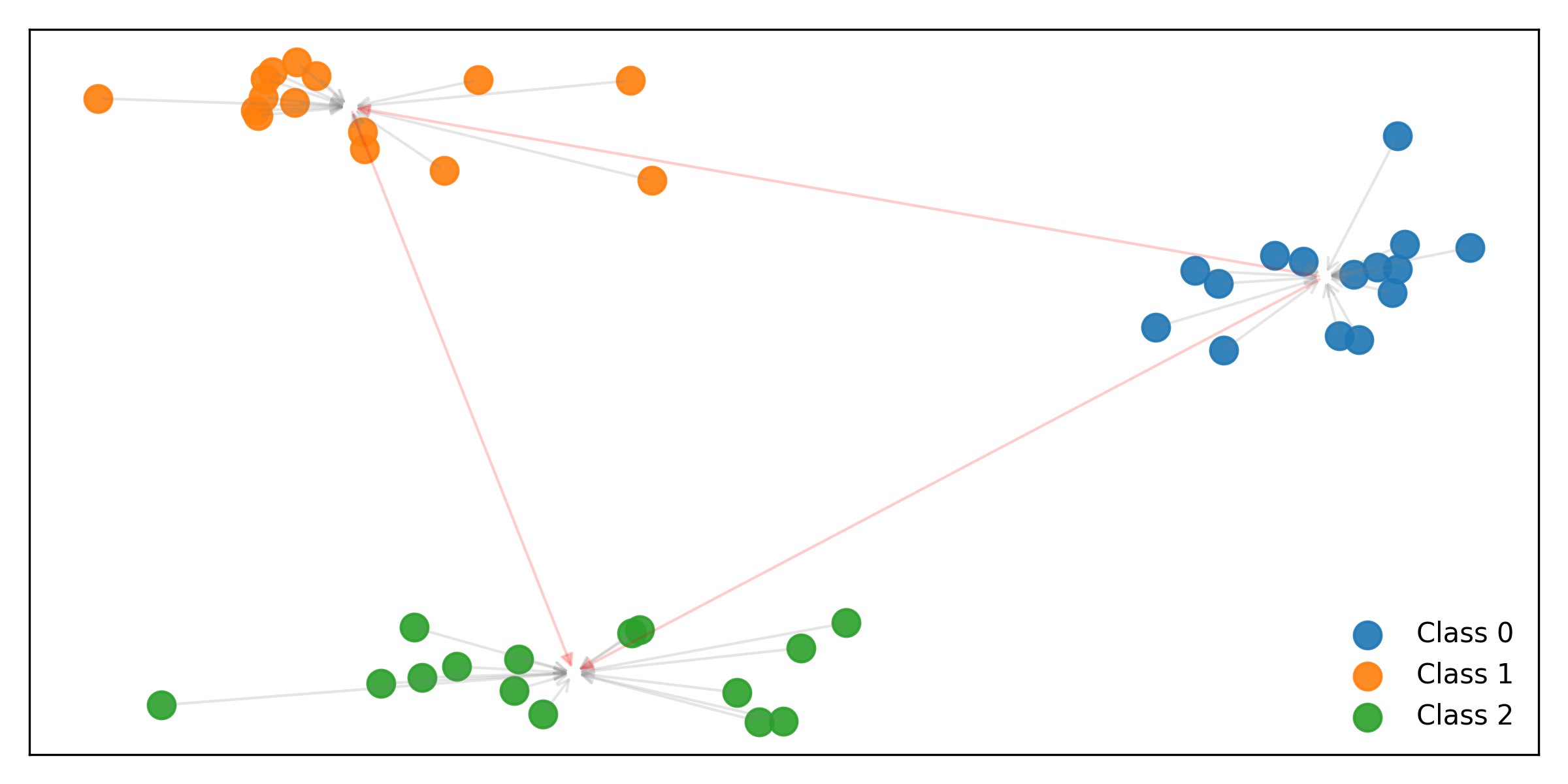}
        \caption{Supervised Contrastive Learning (Class-aware embeddings).}
        \label{fig:sub3}
    \end{subfigure}

    \caption{Visual comparison of supervised learning, contrastive learning and supervised contrastive learning.}
    \label{fig:overall}
\end{figure}

However, conventional contrastive learning does not explicitly use class labels, only information about whether two (or more) instances belong to the same class. When class labels are available, supervised contrastive learning can be employed to extend the standard contrastive framework. This approach leverages label information to explicitly pull together samples from the same class while pushing apart those from different classes, leading to more compact and discriminative clusters in the representation space. Figure \ref{fig:overall} illustrates the latent space representations generated by the approaches discussed above. 

Figure \ref{fig:sub1} shows the decision boundaries learned by a model trained using standard supervised learning. Figure \ref{fig:sub2} depicts instance pairs that are either pulled together (if they belong to the same class) or pushed apart (if they belong to different classes). Importantly, this approach does not utilize explicit class labels, only information about whether pairs share the same class, thus no global class structure emerges. Figure \ref{fig:sub3} displays the latent space of a model trained with supervised contrastive learning, where instances naturally form well-defined clusters corresponding to their classes.

Since supervised contrastive learning was employed in this work, a technique in which the model learns to generate clustered embeddings, models were allowed to learn their own representations directly. Consequently, there was no need to manually extract handcrafted features or statistical descriptors, as doing so could introduce bias into the learned embeddings and degrade overall performance.

\subsection{System Architecture}
\begin{figure}[H]
    \centering
    \includegraphics[width=0.9\linewidth]{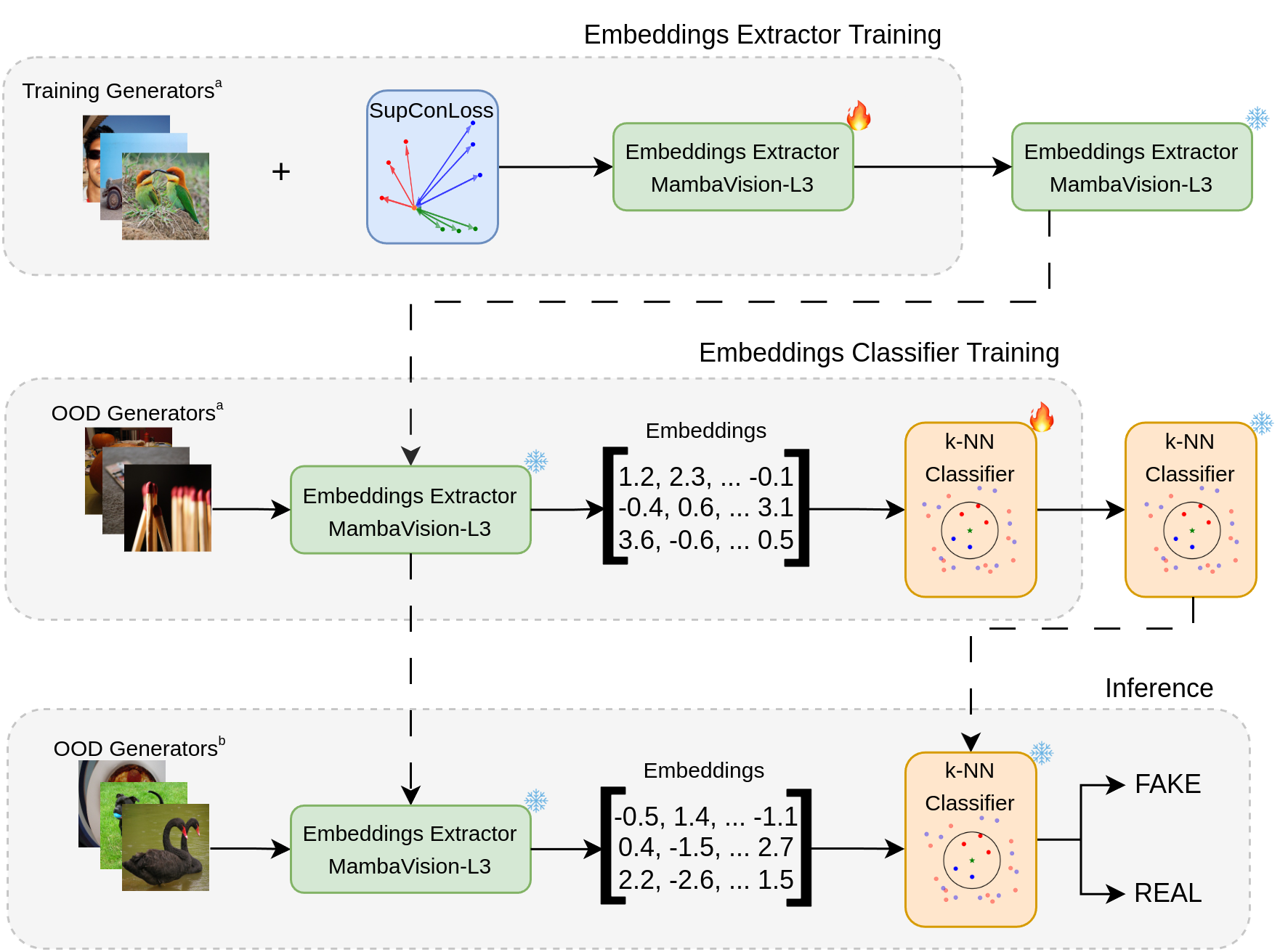}
    \caption{Full architecture of the proposed approach. $^a$ Indicates that training samples were used. $^b$ Denotes that testing samples were used.}
    \label{architecture}
\end{figure}

The base architecture of the proposed model is presented in Figure \ref{architecture}. Two stacked models compose the processing pipeline. The first model is in charge of extracting embeddings of images (embeddings extractor), while the second model classifies those embeddings (embeddings classifier).

\subsubsection{Supervised Contrastive Embedding Learning}

\label{extr}
The embedding extractor employs a MambaVision-L3-256-21K  model \cite{nvidia2024mambavisionl3} as its backbone. This model was selected for its effective integration of long-range context modeling with computational efficiency through selective state-space modeling \cite{hatamizadeh2025mambavision}. Unlike conventional Vision Transformers that rely exclusively on self-attention mechanisms with quadratic complexity, MambaVision achieves linear-time complexity while maintaining strong representational capacity. This architectural design enables scalable feature extraction from high-resolution inputs with reduced memory overhead, a critical requirement for processing large-scale datasets as in this work. Additionally, the model's pretraining on ImageNet-21K provides robust general-purpose visual representations suitable for transfer learning to fake image detection and source attribution. Other families of models were tested, such us EfficientNets \cite{efficientnet}, but they underperformed compared to MambaVision. In order to train the MambaVison model, as commented in previous sections, a supervised contrastive learning approach was employed, hence SupConLoss \cite{khosla2020supervised} was used as loss. 

Let $\mathbf{z}_i \in \mathbb{R}^d$ denote the embedding representation of the $i$-th instance within a batch of $N$ samples, where $d$ denotes the dimensionality of the latent feature space determined by the model’s embedding layer. For each anchor embedding $\mathbf{z}_i$, let $P(i)$ represent the set of indices corresponding to positive samples that share the same class as $\mathbf{z}_i$, and let $A(i)$ denote the set of indices of all other samples in the batch except $\mathbf{z}_i$ itself. The similarity between two embeddings $\mathbf{z}_i$ and $\mathbf{z}_j$ is measured by their dot product $\mathbf{z}_i \cdot \mathbf{z}_j$, which corresponds to the cosine similarity when the embeddings are $\ell_2$-normalized. The temperature parameter $\tau \in \mathbb{R}^{+}$ controls the concentration level of the similarity distribution and can be treated as a learnable scalar. The supervised contrastive loss is then defined as shown in Equation~\ref{supconloss}:

\begin{equation} \label{supconloss} \mathcal{L} = \frac{-1}{N} \sum_{i=1}^{N} \frac{1}{|P(i)|} \sum_{p \in P(i)} \log \frac{\exp(\mathbf{z}_i \cdot \mathbf{z}_p / \tau)} {\sum_{a \in A(i)} \exp(\mathbf{z}_i \cdot \mathbf{z}_a / \tau)} \end{equation}

The loss encourages the model to pull positive pairs closer together in the latent space while pushing negative pairs apart. This setup allows the model to learn a clustered latent space, where embeddings can then be effectively classified.

Specifically, the per-anchor loss computes the log-probability of each positive sample relative to all other samples, scaled by a temperature parameter $\tau$ which controls the concentration of the distribution. By averaging over all anchors in the batch, the total loss ensures that the learned embeddings preserve class-level similarity across the dataset.

\subsubsection{Few-Shot Non-Parametric Classification}
\label{cls}
On the other hand, a k-Nearest Neighbors (k-NN) is employed as the embeddings classifier, which operates without an explicit training phase. Unlike parametric models that learn decision boundaries during training, k-NN is a non-parametric, instance-based learning algorithm that defers all computational operations to inference time. The classifier maintains a reference knowledge base comprising labeled instances, which are used to make predictions through similarity comparisons.
As k-NN follows a supervised learning paradigm, it requires representative examples from each target class to perform classification effectively. To construct this knowledge base, a few-shot learning strategy was adopted. Specifically, a small number of instances are sampled from each class within the testing set, which consists of images created by the testing generators. It is important to note that the testing generator set comprises both generators that were used during the training of the embedding extractor model and generators that were completely unseen during training, enabling evaluation of both in-domain performance and generalization capabilities. These instances are then processed through the already-trained embedding extractor to obtain their corresponding feature representations in the latent space. The resulting embeddings, along with their associated class labels, constitute the reference set for the k-NN classifier (see Figure \ref{architecture}). 

During inference, new instances are classified by first computing their embeddings with the embeddings extractor model (see Section \ref{extr}), and then passed to the embeddings classifier model (see Section \ref{cls}), which computes their similarity to the stored embeddings and assigning labels based on the nearest neighbors in the latent space (see Figure \ref{architecture}).

\subsection{Experimental setup}
Both the embeddings extractor training and inference were performed using two NVIDIA RTX 4500 Ada Generation with 24 GB of VRAM, enabling efficient handling of large datasets and computationally intensive models. All training and evaluation procedures were executed on a Unix-based operating system. The k-NN algorithm, however, was executed exclusively on the CPU, specifically an Intel(R) Xeon(R) Silver 4310 CPU @ 2.10GHz with 24 cores (2 threads per core), without GPU acceleration.

\subsubsection{Datasets}
\label{datasets}
Regarding the datasets, two publicly available benchmarks were employed: GenImage \cite{cite14} and ForenSynths \cite{wang2020cnn}. The GenImage dataset comprises synthetic images produced by eight distinct generative models (see Figure \ref{figure:1}). For each generator, 162,000 generated images are provided for training and 6,000 for testing, with the exception of Stable Diffusion 1.5, which includes 8,000 test samples. A corresponding set of real images is also included for each generator, counting 162,000 for training and 6,000 for testing (8,000 for Stable Diffusion 1.5). Importantly, the synthetic data is curated to follow the ImageNet class distribution \cite{deng2009imagenet}, thereby ensuring content consistency and enabling direct comparisons between real and generated imagery.

In contrast to GenImage, the ForenSynths dataset adopts a different evaluation paradigm. Its training set contains synthetic images produced by a single generator (ProGAN), consisting of 360,000 fake and 360,000 real samples. In the test set, however, images are generated by 12 different synthesis models to assess generalization across unseen generators. In line with prior work in deepfake and synthetic image detection, the testing split from ForenSynths was not utilized in this study. Instead, models trained on the ForenSynths training set were evaluated using the GenImage test set, allowing for a rigorous and fair comparison with existing state-of-the-art methods under a challenging cross-model generalization scenario. 

For the source attribution task, which inherently poses a more complex problem, models were trained using a selected subset of generators from GenImage and evaluated on previously unseen generators using a few-shot learning scheme (see Section \ref{experiments}). This setup better reflects real-world forensic conditions, where newly emerging generative models may not be present in the training data.

Regarding data augmentation, it was deemed unnecessary since both datasets contained a sufficiently large number of samples per class. Each image generator within the datasets produced outputs at different native resolutions, therefore, all images were resized to a uniform resolution of 
$256\times256$ pixels (see Section~\ref{ablations}). The only additional preprocessing step applied was pixel-value normalization to the interval $[0,1]$. Specifically, as all images were in RGB format with pixel values originally constrained to the range $[0,255]$, normalization was performed by dividing all pixel values by $255$.

\subsubsection{Experiments}
\label{experiments}
For the fake image detection task, a single embedding extractor was trained. As described in Section \ref{datasets}, this model was trained using the ForenSynths training set, which contains real images and fake images generated exclusively by ProGAN. Evaluation was conducted using the GenImage test set, ensuring that the model was exposed only to generators unseen during the training phase. The model trained under this configuration is referred to as ESB1.

For the source attribution task, multiple embedding extractors were trained. Each model was trained using real images and synthetic images from three generative models selected from GenImage (see Table \ref{data} for the generator groupings). The rationale behind the selection of each experimental setup is as follows:

\begin{itemize}
\item The selected generators in ES1 are based on diverse architectural designs. This setup enables an evaluation of how a classifier trained on data from multiple generative architectures performs, providing insights into the model’s ability to generalize across heterogeneous generator types.
\item ES2 aims to investigate the temporal generalization capability of the classifier. Specifically, by training on images produced by an earlier version of a generator (SD\_1.4) and evaluating on a subsequent version (SD\_1.5), ES2 assesses whether the model can effectively detect newer generations of synthetic images originating from the same source.

\item ES3 examines whether combining two consecutive versions of the same generator (SD\_1.4 and SD\_1.5) during training leads to improved classification performance. This setup explores the benefits of exposure to intra-generator variability for enhancing robustness and discriminative power.

\item ES4 evaluates the impact of incorporating images from proprietary generators (e.g., Midjourney) in the training process. ES4 aims to determine whether exposure to such models—often characterized by distinctive generative styles—can further enhance the classifier’s overall performance.

\end{itemize}

All models were trained with real images, as real images are easy and feasible to gather, and can provide valuable information to the model to extract useful features for their classification. The repository with the full code to conduct these experiments is available at \url{https://github.com/JaimeAlvarez18/SupConLoss_fake_image_detection}.

The same testing protocol was followed for all models: test data included real images and synthetic images from all generators in GenImage. This setup enables performance assessment on both seen and unseen generators, thereby providing a realistic evaluation of generative source attribution.

Given the numerous possible combinations of three generators, four representative subsets were constructed (denoted ES1–ES4 in Table \ref{data}), each corresponding to a unique trio of generators and real images. Each subset was used to train a distinct embedding extractor. Additionally, one more embedding extractor (ES5) was trained using all available generators in GenImage. This model serves as an upper-bound performance reference, whereas ESB1 serves as a lower-bound benchmark, since it was trained without data from any generator appearing in the GenImage test set.

Across all experiments, 150 images per generator were used for the few-shot classification stage. This quantity represents a practical and realistic amount of data that could reasonably be collected for a newly emerging generative model in real forensic scenarios. Additionally, as discussed in Section \ref{knn}, a complementary study was conducted to analyze the sensitivity of the model’s performance to this hyperparameter, demonstrating how variations in the number of available samples influence attribution accuracy.

\begin{table}[H]

    \centering
    \caption{Usage of the distinct generators to train the embeddings extractor models for the source attribution problem.}
    \begin{tabular}{c:ccccccccc}
    
    \label{data}
\textbf{Experiment} & \textbf{ADM} & \textbf{SD\_1.4} & \textbf{SD\_1.5} & \textbf{VQDM} & \textbf{Midjourney} & \textbf{Real} & \textbf{Glide} & \textbf{BigGan} & \textbf{Wukong} \\ \hline

         ES1& \textcolor{green}{\ding{51}} & \textcolor{red}{\ding{55}} & \textcolor{red}{\ding{55}} & \textcolor{green}{\ding{51}} & \textcolor{red}{\ding{55}} & \textcolor{green}{\ding{51}} & \textcolor{red}{\ding{55}} & \textcolor{green}{\ding{51}} & \textcolor{red}{\ding{55}}\\
         ES2& \textcolor{red}{\ding{55}} & \textcolor{green}{\ding{51}} & \textcolor{red}{\ding{55}} & \textcolor{red}{\ding{55}} & \textcolor{red}{\ding{55}} & \textcolor{green}{\ding{51}} & \textcolor{green}{\ding{51}} & \textcolor{red}{\ding{55}} &\textcolor{green}{\ding{51}} \\
         ES3& \textcolor{red}{\ding{55}} & \textcolor{green}{\ding{51}} & \textcolor{green}{\ding{51}} & \textcolor{red}{\ding{55}} & \textcolor{green}{\ding{51}} & \textcolor{green}{\ding{51}} & \textcolor{red}{\ding{55}} & \textcolor{red}{\ding{55}} & \textcolor{red}{\ding{55}}\\
         
         ES4& \textcolor{green}{\ding{51}} & \textcolor{red}{\ding{55}} & \textcolor{red}{\ding{55}} & \textcolor{green}{\ding{51}} & \textcolor{green}{\ding{51}} & \textcolor{green}{\ding{51}} & \textcolor{red}{\ding{55}} & \textcolor{red}{\ding{55}} & \textcolor{red}{\ding{55}}\\
         ES5& \textcolor{green}{\ding{51}} & \textcolor{green}{\ding{51}} & \textcolor{green}{\ding{51}} & \textcolor{green}{\ding{51}} & \textcolor{green}{\ding{51}} & \textcolor{green}{\ding{51}} & \textcolor{green}{\ding{51}} & \textcolor{green}{\ding{51}} & \textcolor{green}{\ding{51}}\\
         ESB1& \textcolor{red}{\ding{55}} & \textcolor{red}{\ding{55}} & \textcolor{red}{\textcolor{red}{\ding{55}}} & \textcolor{red}{\ding{55}} & \textcolor{red}{\ding{55}} & \textcolor{green}{\ding{51}} & \textcolor{red}{\ding{55}} & \textcolor{red}{\ding{55}} & \textcolor{red}{\ding{55}}\\
    \end{tabular}
\end{table}

\subsection{Ablation Studies}
\label{ablations}
Supervised contrastive learning evaluates each image relative to all other images within a training batch (see Equation \ref{supconloss}). This formulation enables the model to simultaneously attract samples belonging to the same class while repelling samples from different classes within a single optimization step. Positive pairs (images sharing the same class label) promote compact intra-class clusters, whereas negative pairs encourage inter-class separation by pushing apart embeddings from different classes \cite{chen2020simple}. As a result, the number of negatives available in each batch directly influences the quality of contrastive representation learning.

Since the quantity of negative comparisons increases with batch size, larger batches typically yield stronger cluster separation and improved downstream performance \cite{chen2022wef}. Motivated by this observation, a batch size of 6,000 was selected for models ES1–ES5 and 4,000 for ESB1. Alternative batch sizes could lead to different degrees of feature separation and, consequently, different classification outcomes.

Another important hyperparameter is the dimensionality of the embedding space. In this work, embeddings are represented as 1,000-dimensional vectors, corresponding to the output of the MambaVision-L3 encoder. While larger embedding spaces allow the model to encode richer semantic structures, they also introduce higher memory consumption and computational cost, which constrained further scaling.

All input images were resized to 256$\times$256 pixels. Higher resolutions were not feasible due to GPU memory limitations in combination with the large batch sizes adopted. Given this trade-off, batch size was prioritised over image resolution, as maximizing the number of negative pairs was deemed more critical for representation quality.

Finally, the downstream k-NN classifier was configured with 
k=11. A series of experiments was conducted to validate this setting; however, detailed analysis of k sensitivity lies beyond the scope of this work.

\subsection{Open Set Attribution}
Given the difficulty of the strict source attribution task, several authors have instead adopted open-set recognition approaches\cite{vaze2021open,yang2023progressive} to address this problem . In this paradigm, all images originating from unseen generators are grouped into a single “unknown” class. The classification model is therefore required only to determine whether an input image belongs to one of the generators encountered during training or should be assigned to the unknown class. This formulation substantially simplifies the task, as the model no longer needs to identify the specific generator of images from previously unseen sources, but merely to detect that they do not correspond to any of the known classes.

To evaluate the classification approach, a series of data splits was performed. The splits followed the same protocol as in \cite{yang2022deepfake, bui2022repmix, yang2023progressive, cioni2024clip}, ensuring that the results are directly comparable. Specifically, four splits were generated, metrics were computed for each split, and the final performance was obtained by averaging the results across all splits.

The quality of the trained models under different data splits was evaluated using Accuracy (Acc), Area Under the ROC Curve (AUC), and the Open-Set Classification Rate (OSCR) \cite{dhamija2018reducing}. Accuracy was computed only on images generated by the models included in the training set, whereas AUC and OSCR were calculated using images from both seen and unseen generators. This evaluation strategy follows the methodology adopted in prior work, enabling assessment of model behavior in both seen and unseen generators.

The OSCR metric is defined using both the Correct Classification Rate (CCR) and the False Positive Rate (FPR). The CCR is the fraction of examples from the seen data \(D_S\) for which the correct class \(\hat{k}\) has the maximum predicted probability and this 
probability exceeds a threshold \(\tau\) (see Equation \ref{CCR}). The FPR is defined as the fraction of samples from the unknown data \(D_U\) that are classified as any known class \(k\) with a predicted probability greater than the threshold \(\tau\) (see Equation \ref{FPR}. The OSCR is then computed as the area under the curve obtained by plotting the CCR against the FPR.
\begin{equation}
\label{CCR}
\mathrm{CCR}(\tau) = 
\frac{
\left| \left\{ x \mid x \in \mathcal{D}_S \land \arg\max_k P(k \mid x) = \hat{k} \land P(\hat{k} \mid x) > \tau \right\} \right|
}{
|\mathcal{D}_S|}
\end{equation}

\begin{equation}
\label{FPR}
\mathrm{FPR}(\tau) =
\frac{
\left| \left\{ x \mid x \in \mathcal{D}_U \land \max_k P(k \mid x) \ge \tau \right\} \right|
}{
|\mathcal{D}_U|}
\end{equation}

\section{Experimental Results and Discussion}
\label{results}

\subsection{Detection Performance on Synthetic Images}

Table \ref{binary} summarizes the accuracy achieved by the proposed method compared to state-of-the-art approaches. Importantly, all models listed in the table were trained using identical training data, which ensures a fair and direct comparison.

Overall, the proposed model consistently outperforms the competing methods across most tested generators. Although the approach in \cite{cite16} slightly surpasses our model for 4 out of the 8 generators, its performance is markedly unstable across different generative sources. In contrast, our method demonstrates strong robustness and stability, which is essential in practical forensic scenarios. Specifically, the accuracy of our model ranges from 86.8\% (VQDM) to 96.5\% (BigGAN), indicating reliable detection capabilities even when dealing with diverse and previously unseen synthesis techniques.

Conversely, \cite{cite16} exhibits a pronounced variance in performance, with reported accuracies fluctuating between 55.4\% and 98.0\%. This variability highlights a key limitation in current approaches: strong performance on specific generators does not necessarily translate to effective generalization. Such behavior becomes particularly problematic as new generative models continue to emerge rapidly.

Moreover, the proposed model achieves a 5.2\% higher average accuracy, driven by substantial improvements on generators such as Midjourney and Wukong, where \cite{cite16} underperforms significantly. This reflects the enhanced ability of our feature representations to capture synthesis artifacts that transfer across generator architectures.

In the context of fake image detection, where real-world deployment conditions demand resilience to unknown or evolving manipulations, a model that delivers consistent high performance across a wide set of generators is inherently more valuable than one that excels only in limited cases. The improved generalization capabilities demonstrated by our approach therefore represent an important advancement toward robust synthetic image forensics.
\begin{table}[H]
    \caption{Comparison of accuracy between SotA models and ours in the AI-generated problem.}
    \centering
    \begin{tabular}{c:cccccccc:c}
    \label{binary}
         \textbf{Model} & \textbf{ADM} & \textbf{BigGan} & \textbf{Glide} & \textbf{Midjourney} & \textbf{VQDM} & \textbf{SD 1.4} & \textbf{SD 1.5} & \textbf{Wukong} & \textbf{Average} \\ \hline
        \cite{uni} & $68.1$ & $95.3$ & $64.0$ & $57.4$ & $85.2$ & $61.2$ & $63.0$ & $71.3$ & $70.7$ \\ 
        \cite{cite15} & $76.4$ & $67.1$ & $72.4$ & $58.4$ & $54.4$ & $49.6$ & $49.8$ & $55.4$ & $60.4$ \\
        \cite{cite16} & $87.1$ & $94.2$ & $\textbf{96.3}$ & $65.1$ & $\textbf{95.6}$ & $\textbf{98.0}$ & $\textbf{97.1}$ & $55.4$ & $86.1$ \\
        \cite{grad} & $61.4$ & $82.1$ & $70.8$ & $67.4$ & $67.8$ & $63.0$ & $64.2$ & $70.8$ & $68.4$ \\
        ESB1 (ours) & $\textbf{87.5}$ & $\textbf{96.5}$ & $87.7$ & $\textbf{91.6}$ & $86.8$ & $94.4$ & $94.1$ & $\textbf{91.6}$ & $\textbf{91.3}$ \\
    \end{tabular}
\end{table}

Table~\ref{aucs1} presents a comparative analysis between the proposed model and several state-of-the-art approaches, using the Area Under the Curve (AUC) metric. These results further reinforce the earlier claim that the proposed model demonstrates superior robustness. Not only does it outperform all competing methods across nearly all evaluated generators, with the exception of GLIDE, but it also exhibits the lowest performance variance across generators. This reduced variance indicates that the model generalizes more effectively to unseen generators during the training phase. Moreover, the proposed method achieves the highest average AUC across all generators, surpassing previous state-of-the-art results by 3.1\%.
\begin{table}[H]
    \caption{Comparison of AUC between SotA models and ours in the AI-generated problem.}
    \centering
    \begin{tabular}{c:cccccc:c}
         \textbf{Model} & \textbf{ADM} & \textbf{BigGan} & \textbf{Glide} & \textbf{Midjourney}  & \textbf{SD 1.4} & \textbf{SD 1.5} & \textbf{Average} \\ \hline
        \cite{cite12} & $79.9$  & $89.3$ & $\textbf{99.7}$ & $81.7$ & $91.3$ & $91.3$ & $88.9$ \\ 
        \cite{npr} & $86.3$  & $87.5$ & $79.3$ & $77$ & $64.5$ & $64.5$ & $76.5$ \\ 
        \cite{uni} & $86.7$  & $94.5$ & $80.8$ & $66.2$ & $89.5$ & $89.5$ & $84.5$ \\ 
        \cite{cite20} & $82.5$  & $88.1$ & $76.5$ & $40.7$ & $77.4$ & $77.4$ & $73.8$ \\ 
        ESB1 (ours) & $\textbf{87.5}$ & $\textbf{96.5}$ & $87.7$ & $\textbf{91.6}$ & $\textbf{94.4}$ & $\textbf{94.1}$ & $\textbf{92.0}$ \\
    \end{tabular}
    \label{aucs1}
\end{table}

\subsection{Generator Source Attribution Results}

\label{source}
In the source attribution problem, more classes were available for the model in the training set, hence better results were obtained leveraging the advantages of SupConLoss. For each model, precision, recall, and F1-score were computed separately for the classes included during training and those excluded from training, as shown in Table  \ref{resultss}.This approach allows for a direct evaluation of model  performance on both seen and unseen classes. Table \ref{resultss} summarizes these metrics, highlighting the differences in performance between unseen and seen generators.
\begin{table}[H]
    \caption{Results of the trained models over unseen and seen generators.}
    \centering
    \begin{tabular}{c:cccccc}
     \label{resultss}

             & \multicolumn{3}{c}{\textbf{Unseen}} & \multicolumn{3}{c}{\textbf{Seen}}\\
             \textbf{Model} & \textbf{Precision} & \textbf{Recall} & \textbf{F1-Score} & \textbf{Precision} & \textbf{Recall} & \textbf{F1-Score} \\ \hline
            ES1 & $54.0$ & $51.3$ & $52.3$ & $96.3$ & $98.3$ & $97.3$ \\
            ES2 & $65.9$ & $62.6$ & $63.6$ & $68.4$ & $69.8$ & $68.9$ \\
            ES3 & $39.2$ & $44.9$ & $36.7$ & $60.3$ & $47.7$ & $44.3$ \\
            ES4 & $55.1$ & $50.0$ & $50.7$ & $91.6$ & $97.2$ & $94.2$ \\
            ES5 & - & - & - & $80.5$ & $80.3$ & $80.4$ \\
            ESB1 & $41.5$ & $33.3$ & $32.43$ & $65.8$ & $67.4$ & $66.6$ \\
    \end{tabular}
\end{table}

Among the models trained explicity for the source attribution task, ES1 and ES4 achieved the best overall performance, considering both seen and unseen generators. This outcome highlights the importance of generator selection in shaping the discriminative power of the learned embeddings. Not all generators provide equally informative cues for training; their contribution depends on the diversity and distinctiveness of the artifacts they introduce.

Notably, ES3 obtained the lowest results across both evaluation settings (apart from the ESB1 baseline), suggesting that the generators used in its training set were less informative or offered more homogeneous characteristics, limiting the model’s ability to learn robust and transferable features.

The ES1 model achieved the highest F1-scores, reaching 97.3\% on seen generators and 52.3\% on unseen ones. ES4 closely followed, with 94.2\% and 50.7\%, respectively. Interestingly, ES1 and ES4 share all training generators except one, implying that the common set (BigGAN, VQDM, and ADM) provides the most representative and informative features for generator attribution. In contrast, ES2 and ES3, which share Stable Diffusion 1.4 (SD\_1.4) as a training generator, exhibit comparatively weaker results, indicating that this generator contributes less distinctive cues for the model to learn from.

These findings underline that the choice of training generators plays a critical role in shaping model generalization, and that some generative models inherently contain richer forensic signatures that are more transferable across unseen synthesis techniques.

Table \ref{f1} reports the F1-scores obtained for each unseen generator, allowing us to identify which generative models pose a greater challenge for attribution. Notably, ADM achieves high F1-scores even in ES2 and ES3 (the two weaker-performing models overall), indicating that ADM exhibits more distinctive and transferable generative fingerprints, making it comparatively easier for models to generalize to.

Similarly, ES1 and ES4 achieve strong performance on Glide, Midjourney, and BigGAN, suggesting that these generators introduce characteristic synthesis artifacts that align well with the feature space learned during training. These results reinforce the idea that the choice of training generators significantly influences the model’s ability to cluster unseen generators correctly.

Conversely, Stable Diffusion 1.4, Stable Diffusion 1.5, and Wukong consistently lead to the lowest attribution performance across all models. This suggests that their underlying generative mechanisms produce visual patterns or artifacts that are less correlated with those present in the training data. In other words, models trained on other generators struggle to generalize toward these architectures, possibly due to higher diversity, more advanced denoising strategies, or artifact suppression techniques present in diffusion-based models.

\begin{table}[H]
    \caption{Comparison of F1 scores obtained by each model over unseen generators.}
    \centering
    \begin{tabular}{c:ccccccccc}
    \label{f1}
\textbf{Model} & \textbf{ADM} & \textbf{SD 1.4} & \textbf{SD 1.5} & \textbf{VQDM} & \textbf{Midjourney} & \textbf{Real} & \textbf{Glide} & \textbf{BigGan} & \textbf{Wukong} \\ \hline

ES1 & - & $32.3$ & $36.2$ & - & $63.3$ & - & $95.0$ & - & $39.9$ \\
ES2 & $71.0$ & - & $43.2$ & $63.6$ & $63.9$ & - & - & $88.5$ & - \\
ES3 & $53.6$ & - & - & $27.3$ & - & - & $39.7$ & $46.0$ & $16.8$ \\
ES4 & - & $39.3$ & $28.4$ & - & - & - & $77.7$ & $87.8$ & $28.0$ \\
ES5 & - & - & - & - & - & - & - & - & - \\
ESB1 & $34.4$ & $25.4$ & $30.2$ & $10.1$ & $26.1$ & - & $32.9$ & $88.1$ & $13.1$ \\
    \end{tabular}
\end{table}

Remarkably, ESB1 obtained substantially poorer results in the source attribution task compared to ES1–ES5, despite its strong performance on the fake image detection task. This discrepancy arises from the fact that ESB1 was trained exclusively on real images and images generated by ProGAN, meaning that all generators evaluated in the source attribution experiments were unseen for this model. In contrast, ES1–ES4 were trained on real images alongside samples from three different generators, enabling the use of SupConLoss to significantly strengthen their ability to discriminate between generator-specific fingerprints. ES5, trained on real images and samples from all evaluated generators, unsurprisingly achieved the best performance (see Table~\ref{resultss}).

Overall, the source attribution task is inherently more challenging than fake image detection, as it requires models to distinguish between subtle, generator-specific fingerprints rather than simply determining whether an image is real or synthetic. This increased complexity explains the larger performance gap between models with limited generator exposure and those trained on broader generator sets.

\subsection{Sensitivity Analysis on Few-Shot instances}
\label{knn}
\begin{figure}[H]
    \centering
    \includegraphics[width=0.8\linewidth]{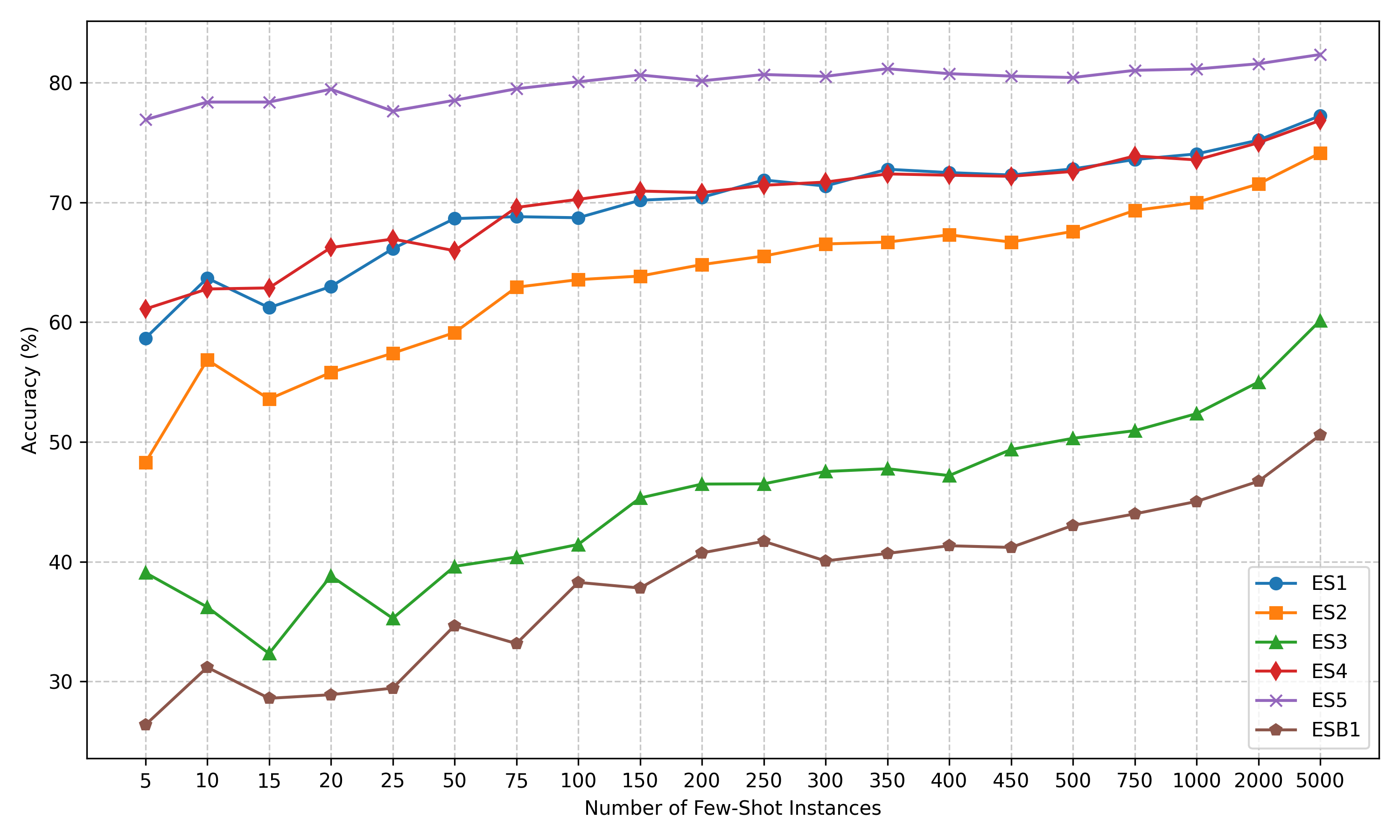}
    \caption{Accuracies obtained in the source attribution task for all trained models, evaluated under different numbers of few-shot samples per generator. }
    \label{fig_knn}
\end{figure}

All the aforementioned results were obtained using just 150 instances per generator, in a few-shot approach. Using 150 images per generator is feasible, however, a study was conducted varying this number to assess how the metrics vary with respect to this hyperparameter. Figure \ref{fig_knn} depicts the metrics of all models trained in this work (ES1-ES5 and ESB1) for the source attribution problem modifying the number of instances per generator as base knowledge for the k-NN.

As expected, increasing the number of few-shot instances per generator consistently improves source attribution performance. When the number of samples is very small (fewer than 50 images), performance exhibits substantial variability, largely due to the randomness of the sample selection and the resulting bias introduced by insufficient class representation. Although the metrics gradually increase as more instances become available, the magnitude of improvement is relatively modest for most models. For example, ES1 and ES4 show only a ~7\% gain in accuracy when increasing from 150 to 5,000 samples per unseen generator.

This improvement becomes even less pronounced for ES5, which was trained using all available generators. For this model, the same increase in few-shot instances results in only a ~2\% gain, reflecting that ES5 already captures highly informative latent representations and therefore it has limited room for improvement through additional retrieval samples.

In contrast, ESB1, which was trained solely on ProGAN and thus did not encounter any of the test generators during training, benefits significantly from the additional data: its accuracy increases by over 12\% when scaling from 150 to 5,000 samples. This behavior is coherent with expectations, models with weaker initial generalization capabilities (such as ESB1) gain more from additional few-shot support examples, whereas stronger and more fully trained models (such as ES5) exhibit diminishing returns.

These findings confirm that while larger few-shot support sets can improve attribution performance, a moderate sample size (150 images per generator) already achieves a favorable trade-off between performance and data acquisition effort, which is crucial for real-world forensic deployment scenarios.

\subsection{Open Set Attribution Evaluation}
The results for this task are presented in Table \ref{tab:results}, which reports the metrics for all splits as well as their average. For all splits except the first, the accuracy is nearly 100\%, indicating that the models can effectively classify images from generators seen during training. The lower performance on Split-1 causes the proposed method to fall short by 0.49\%. This variation across splits can be attributed to the differing difficulty of distinguishing images from specific generators, as some leave more distinctive fingerprints than others.

Regarding the open-set metrics, which include images from both seen and unseen generators, the proposed method outperforms other state-of-the-art approaches. In particular, it achieves a 14.7\% improvement in AUC. Moreover, the standard deviation is only 0.53\%, substantially lower than competing methods, indicating superior generalization. For the OSCR metric, a gain of 4.27\% is observed, accompanied by a reduction in standard deviation of 0.72\%, further demonstrating the robustness of the proposed approach in detecting images from unknown generators, a critical requirement in this domain.

\begin{table}[H]
\centering
\caption{Comparison of methods on the source attribution benchmark dataset.}
\begin{tabular}{c:ccc}
\textbf{Model}& \textbf{Closed Acc} & \textbf{Open AUC} & \textbf{Open OSCR} \\\hline
\cite{yang2022deepfake} & $93.83 \pm 7.72$ & $61.27 \pm 6.70$ & $75.08 \pm 11.02$ \\
\cite{bui2022repmix} & $88.98 \pm 4.36$ & $61.93 \pm 4.26$ & $57.92 \pm 1.73$ \\
\cite{yang2023progressive} & $70.00 \pm 25.95$ & $67.00 \pm 6.08$ & $53.35 \pm 19.82$ \\
\cite{cioni2024clip} & $\mathbf{97.82 \pm 2.51}$ & $81.39 \pm 3.28$ & $80.78 \pm 4.02$ \\
Average ours &$97.33 \pm 4.33$ & $\mathbf{96.09 \pm 0.53}$ & $\mathbf{85.05 \pm 3.30}$ \\\hline
Split-1 (ours) &$89.82$&$95.16$&$79.36$\\
Split-2 (ours)&$99.67$&$96.48$&$86.46$\\
Split-3 (ours) &$99.91$&$96.30$&$87.00$\\
Split-4 (ours) &$99.93$&$96.40$&$87.43$\\
\end{tabular}
\label{tab:results}
\end{table}

\subsection{Explainability and Latent Space Analysis}

\begin{figure}[H]
    \centering
    \begin{subfigure}[b]{0.48\textwidth}
        \centering
        \includegraphics[width=\linewidth]{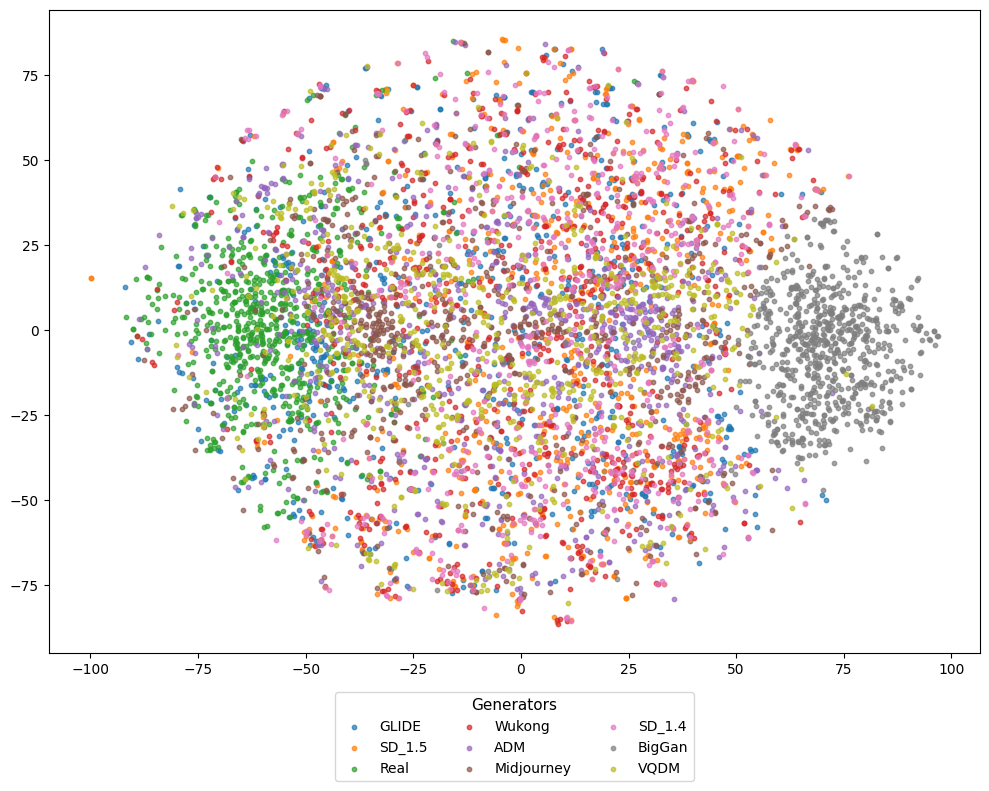}
        \caption{Latent space visualization for the ESB1 model.}
        \label{embs_ESB1}
    \end{subfigure}
    \hfill
    \begin{subfigure}[b]{0.48\textwidth}
        \centering
        \includegraphics[width=\linewidth]{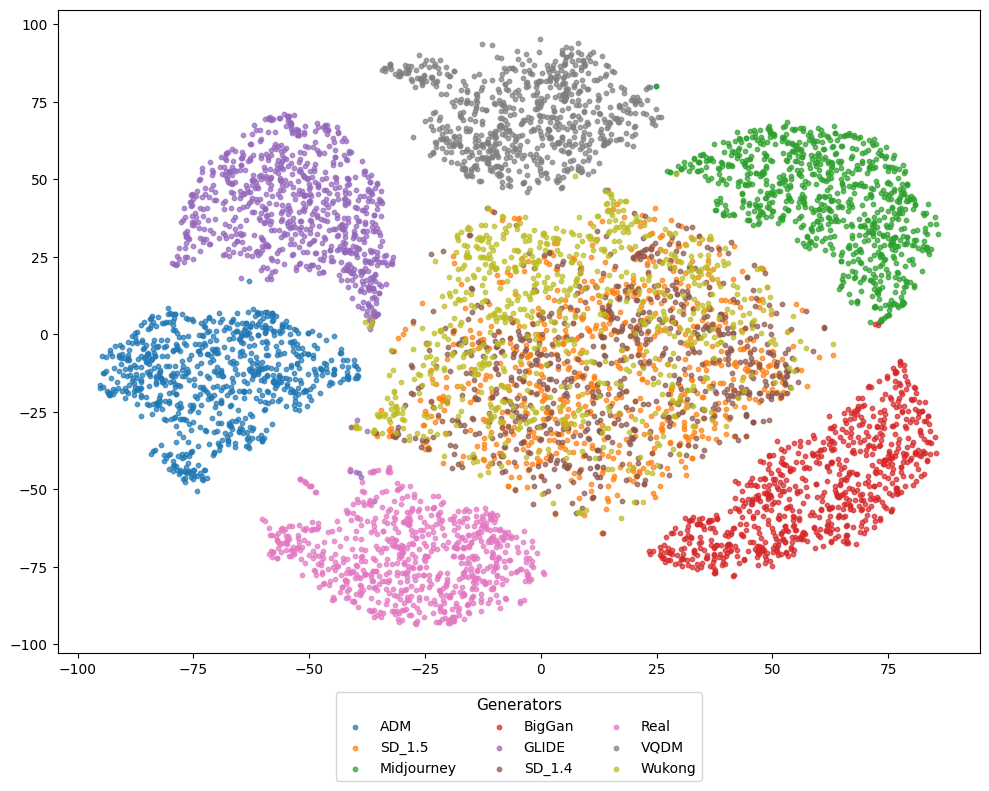}
        \caption{Latent space visualization for the ES5 model.}
        \label{embs_ES5}
    \end{subfigure}
    \hfill
    \caption{Latent space visualizations for the lower-bound (ESB1) and upper-bound (ES5) embedding extractor models.}
    \label{embs}
\end{figure}

To effectively analyze the results obtained from the conducted experiments (see Section \ref{experiments}), the 1,000-dimensional embeddings extracted from the base images were reduced to two dimensions to allow for comprehensive visualization. The t-SNE (t-distributed Stochastic Neighbor Embedding) algorithm was employed for this purpose, as it is a widely used
technique for dimensionality reduction, particularly suited for high-dimensional data such as the one used. Figure \ref{embs_ESB1} depicts the latent space for the model ESB1, while \ref{embs_ES5} shows the latent space for ES5. 
The latent space learned by ESB1 appears disorganized, with no clear clustering structure for most generators. Only real and BigGAN samples form well-defined clusters. This behavior is consistent with the training setup: ESB1 was trained exclusively with ProGAN and real images. Since ProGAN and BigGAN share architectural similarities and produce comparable synthesis artifacts, ESB1 is able to partially cluster BigGAN images despite never being exposed to them during training. Similarly, real images are well represented due to their inclusion in training. However, embeddings from all other unseen generators remain dispersed and overlapping, indicating that ESB1 fails to generalize its learned representation to novel generative sources.

Conversely, the latent space obtained using ES5 exhibits clear and compact clusters for nearly all generators, demonstrating strong inter-class separability and improved generalization. Only three generators (Stable Diffusion 1.4, Stable Diffusion 1.5, and Wukong) do not form distinctly separated clusters. Despite being included in training, their embeddings remain highly overlapping. This suggests that these diffusion-based models produce very similar forensic signatures, consistent with their architectural relationship (SD 1.5 being an incremental evolution of SD 1.4).

\begin{figure}[H]
    \centering
    \begin{subfigure}[b]{0.48\textwidth}
        \centering
        \includegraphics[width=\linewidth]{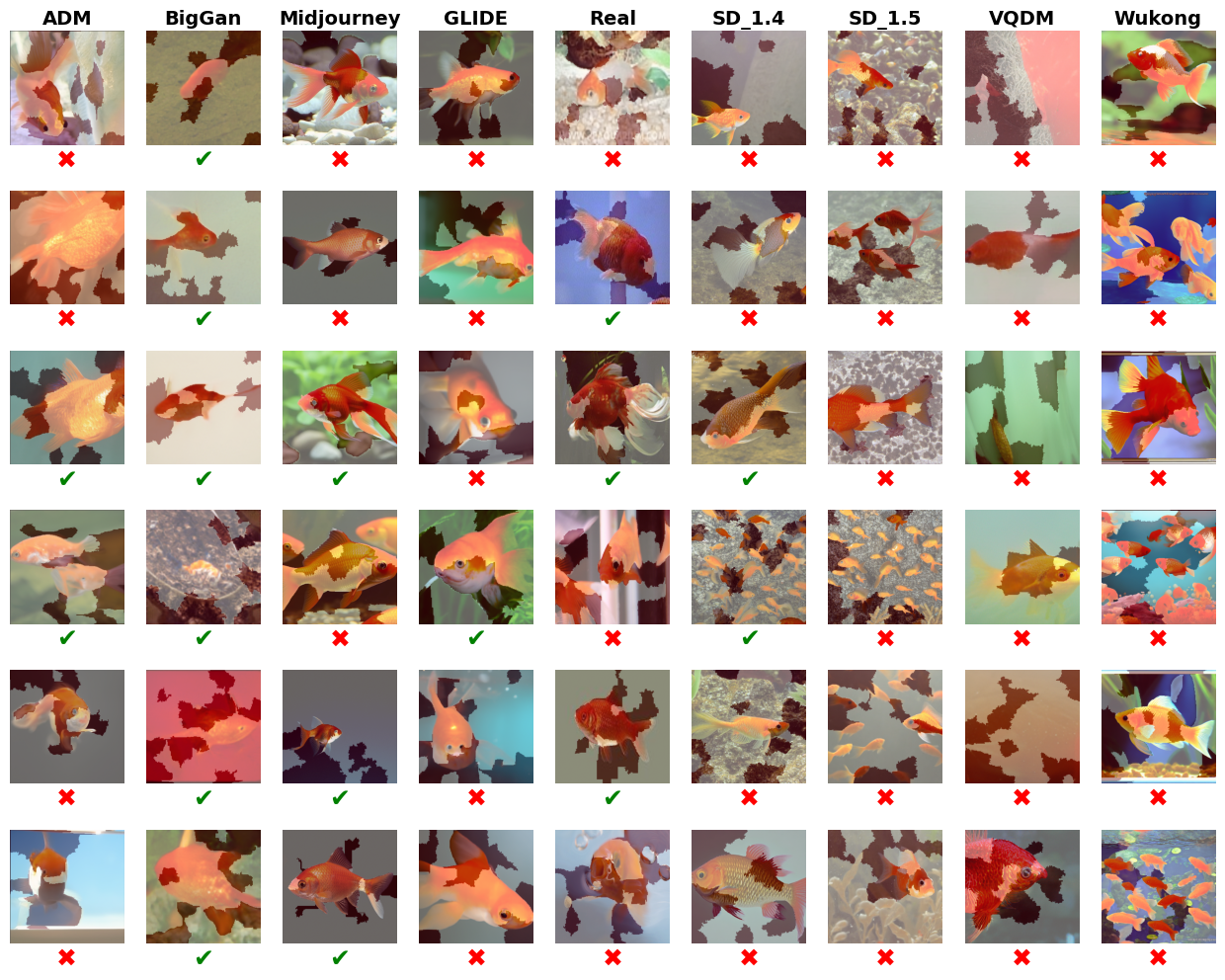}
        \caption{LIME explanation for the ESB1 model.}
        \label{lime_ESB1}
    \end{subfigure}
    \hfill
    \begin{subfigure}[b]{0.48\textwidth}
        \centering
        \includegraphics[width=\linewidth]{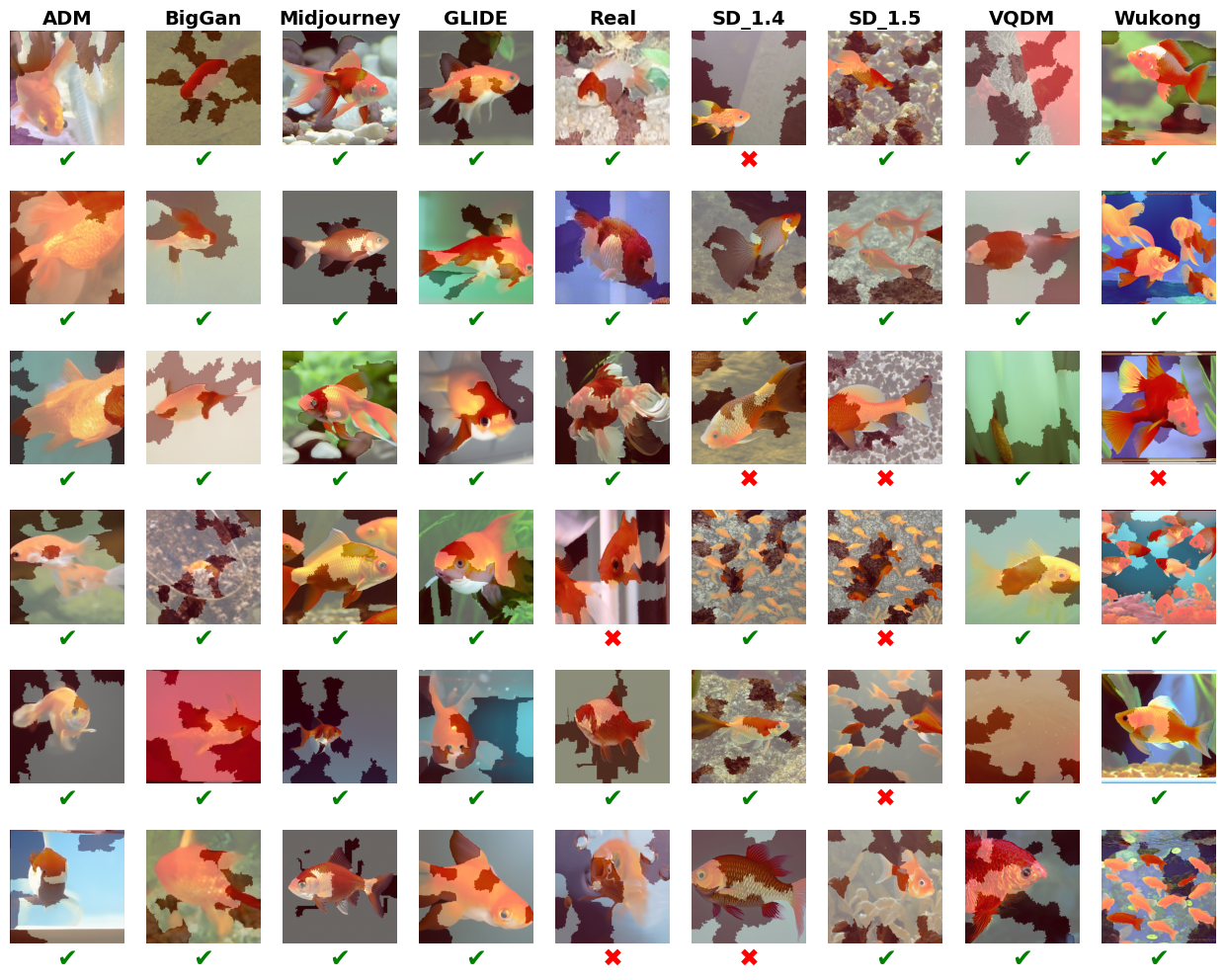}
        \caption{LIME explanation for the ES5 model.}
        \label{lime_ES5}
    \end{subfigure}
    \hfill
    \caption{LIME explanations for the lower-bound (ESB1) and upper-bound (ES5) embedding extractor models.}
    \label{lime}
\end{figure}

Figures \ref{lime_ESB1} and \ref{lime_ES5} depict the output of the Local Interpretable Model-Agnostic Explanations (LIME) method applied to ESB1 and ES5, respectively. The highlighted regions correspond to the patches that contribute most strongly to the embedding assignment for each image. These visualizations provide insight into which areas of the image the model considers informative for distinguishing between different generators'  characteristics.

Remarkably, the ESB1 model fails to correctly classify many instances from most generators except BigGAN. This behavior is expected, as ESB1 was trained exclusively on real images and images generated by ProGAN, a generator architecturally similar to BigGAN. Consequently, ESB1 struggles to generalize to images from other generators. Nevertheless, it manages to correctly classify a subset of samples from ADM, Midjourney, and SD\_1.4. These observations are consistent with the results reported in Section~\ref{source}, where ESB1 (trained with only two classes) was evaluated on a challenging source attribution task involving multiple generators entirely unseen during training.

In contrast, ES5 achieves strong performance across almost all generators, with the exception of SD\_1.4 and SD\_1.5, which aligns with the metrics discussed in Section~\ref{source}. Notably, the model does not rely solely on the appearance of the fish itself, as it also incorporates background cues to determine the generating model. Furthermore, the regions of interest are not fixed, the model’s attention shifts across various parts of the fish, including the beak, fins, and lower body depending on the generator, highlighting its ability to leverage diverse visual features for the source attribution task.

\section{Conclusions and Future Directions}
\label{conclusions}
In this work, we introduced a novel two-stage forensic framework that effectively addresses one of the primary limitations of current synthetic image detection systems: their inability to generalize to newly emerging generative models without computationally costly retraining. By combining supervised contrastive learning for embedding extraction with a non-parametric, few-shot classification strategy, the proposed approach achieves state-of-the-art performance in both fake image detection and generative model source attribution.

The experimental evaluation demonstrated that the system maintains high stability and robustness across unseen generators, even when only a limited number of images per generator are available. These findings confirm that the proposed design is well aligned with real-world digital forensics requirements, where emerging architectures introduce a rapid and continuously evolving threat landscape. Moreover, the achieved improvements in generalization and scalability represent a notable step toward practical forensic deployment.

Regarding the future work, a research worth carrying out is the usage of hard negatives. Hard negatives are negatives pairs that the model confuses greatly (negative pairs whose embeddings are very near in the latent space). Some research states that by training models with lots of hard negatives, the results can be greatly enhanced. A possibility would be to do a warm up
training with both of positive and negative pairs. Once the model understands the task, it is retrained with hard negatives, to encourage the model to separate those conflictive samples.

\section*{Acknowledgments}
This work has been partially supported by the following projects: European Comission under IBERIFIER Plus - Iberian Digital Media Observatory (DIGITAL-2023-DEPLOY- 04-EDMO-HUBS 101158511); by H2020 TMA-MSCA-DN TUAI project "Towards an Understanding of Artificial Intelligence via a transparent, open and explainable perspective" (HORIZON-MSCA-2023-DN-01-01, Grant agreement nº: 101168344); and by Comunidad Autonoma de Madrid, CIRMA-CAM Project (TEC-2024/COM-404).

\bibliographystyle{unsrt}  
\bibliography{references}

@inproceedings{glide,
  title={GLIDE: Towards Photorealistic Image Generation and Editing with Text-Guided Diffusion Models},
  author={Nichol, Alexander Quinn and Dhariwal, Prafulla and Ramesh, Aditya and Shyam, Pranav and Mishkin, Pamela and Mcgrew, Bob and Sutskever, Ilya and Chen, Mark},
  booktitle={International Conference on Machine Learning},
  pages={16784--16804},
  year={2022},
  organization={PMLR},
  doi ={10.48550/arXiv.2112.10741}
}

@misc{nvidia2024mambavisionl3,
  author       = {NVIDIA},
  title        = {MambaVision-L3-256-21K},
  year         = {2024},
  howpublished = {\url{https://huggingface.co/nvidia/MambaVision-L3-256-21K}},
  note         = {Accessed: 2025-10-30},
  publisher    = {Hugging Face},
  license      = {NVIDIA Source Code License-NC}
}

@inproceedings{hatamizadeh2025mambavision,
  title={Mambavision: A hybrid mamba-transformer vision backbone},
  author={Hatamizadeh, Ali and Kautz, Jan},
  booktitle={Proceedings of the Computer Vision and Pattern Recognition Conference},
  pages={25261--25270},
  year={2025}
}

@inproceedings{wufew,
  title={Few-Shot Learner Generalizes Across AI-Generated Image Detection},
  author={Wu, Shiyu and Liu, Jing and Li, Jing and Wang, Yequan},
  booktitle={Forty-second International Conference on Machine Learning},
year={2025}
}

@inproceedings{park2025community,
  title={Community forensics: Using thousands of generators to train fake image detectors},
  author={Park, Jeongsoo and Owens, Andrew},
  booktitle={Proceedings of the Computer Vision and Pattern Recognition Conference},
  pages={8245--8257},
  year={2025}
}

@inproceedings{peng2025crafting,
  title={Crafting Synthetic Realities: Examining Visual Realism and Misinformation Potential of Photorealistic AI-Generated Images},
  author={Peng, Qiyao and Lu, Yingdan and Peng, Yilang and Qian, Sijia and Liu, Xinyi and Shen, Cuihua},
  booktitle={Proceedings of the Extended Abstracts of the CHI Conference on Human Factors in Computing Systems},
  pages={1--12},
  year={2025}
}

@article{aziz2025global,
  title={Global-local image perceptual score (glips): Evaluating photorealistic quality of ai-generated images},
  author={Aziz, Memoona and Rehman, Umair and Danish, Muhammad Umair and Grolinger, Katarina},
  journal={IEEE Transactions on Human-Machine Systems},
  year={2025},
  publisher={IEEE}
}

@article{zheng2024breaking,
  title={Breaking semantic artifacts for generalized ai-generated image detection},
  author={Zheng, Chende and Lin, Chenhao and Zhao, Zhengyu and Wang, Hang and Guo, Xu and Liu, Shuai and Shen, Chao},
  journal={Advances in Neural Information Processing Systems},
  volume={37},
  pages={59570--59596},
  year={2024}
}

@inproceedings{uni,
  title={Towards universal fake image detectors that generalize across generative models},
  author={Ojha, Utkarsh and Li, Yuheng and Lee, Yong Jae},
  booktitle={Proceedings of the IEEE/CVF Conference on Computer Vision and Pattern Recognition},
  pages={24480--24489},
  year={2023},
  doi={10.1109/CVPR52729.2023.02345}
}

@inproceedings{stable_diffusion_1,
  title={High-resolution image synthesis with latent diffusion models},
  author={Rombach, Robin and Blattmann, Andreas and Lorenz, Dominik and Esser, Patrick and Ommer, Bj{\"o}rn},
  booktitle={Proceedings of the IEEE/CVF conference on computer vision and pattern recognition},
  pages={10684--10695},
  year={2022},
  doi={10.48550/arXiv.2105.05233}
}

@inproceedings{vqdm,
  title={Vector quantized diffusion model for text-to-image synthesis},
  author={Gu, Shuyang and Chen, Dong and Bao, Jianmin and Wen, Fang and Zhang, Bo and Chen, Dongdong and Yuan, Lu and Guo, Baining},
  booktitle={Proceedings of the IEEE/CVF conference on computer vision and pattern recognition},
  pages={10696--10706},
  year={2022},
  doi ={10.1109/CVPR52688.2022.01043}
}

@article{wukong,
  title={Wukong: A 100 million large-scale chinese cross-modal pre-training benchmark},
  author={Gu, Jiaxi and Meng, Xiaojun and Lu, Guansong and Hou, Lu and Minzhe, Niu and Liang, Xiaodan and Yao, Lewei and Huang, Runhui and Zhang, Wei and Jiang, Xin and others},
  journal={Advances in Neural Information Processing Systems},
  volume={35},
  pages={26418--26431},
  year={2022},
  doi={10.48550/arXiv.2202.06767}
}

@misc{midjourney,
  author = {Midjourney},
  year = {Accessed: 2025-04-21},
  howpublished = {\url{https://www.midjourney.com/}},
}

@article{adm,
  title={Diffusion models beat gans on image synthesis},
  author={Dhariwal, Prafulla and Nichol, Alexander},
  journal={Advances in neural information processing systems},
  volume={34},
  pages={8780--8794},
  year={2021},
 doi ={10.48550/arXiv.2105.05233}
}

@article{cite3,
  title={The real dangers of generative AI},
  author={Allen, Danielle and Weyl, E Glen},
  journal={Journal of Democracy},
  volume={35},
  number={1},
  pages={147-162},
  year={2024},
  publisher={Johns Hopkins University Press},
  doi={10.1353/jod.2024.a915355}
}

@article{cite4,
  title={Image generation: A review},
  author={Elasri, Mohamed and Elharrouss, Omar and Al-Maadeed, Somaya and Tairi, Hamid},
  journal={Neural Processing Letters},
  volume={54},
  number={5},
  pages={4609--4646},
  year={2022},
  publisher={Springer},
  doi={10.1007/S11063-022-10777-X}
}

@article{cite7,
  title={A conversation on artificial intelligence, chatbots, and plagiarism in higher education},
  author={King, Michael R and ChatGPT},
  journal={Cellular and molecular bioengineering},
  volume={16},
  number={1},
  pages={1--2},
  year={2023},
  publisher={Springer},
  doi={10.1007/s12195-022-00754-8}
}

@article{cite8,
  title={Situating the social issues of image generation models in the model life cycle: a sociotechnical approach},
  author={Katirai, Amelia and Garcia, Noa and Ide, Kazuki and Nakashima, Yuta and Kishimoto, Atsuo},
  journal={AI and Ethics},
  pages={1-18},
  year={2024},
  publisher={Springer},
  doi={10.1007/s43681-024-00517-3}
}

@inproceedings{chen2020simple,
  title={A simple framework for contrastive learning of visual representations},
  author={Chen, Ting and Kornblith, Simon and Norouzi, Mohammad and Hinton, Geoffrey},
  booktitle={International conference on machine learning},
  pages={1597--1607},
  year={2020},
  organization={PmLR}
}

@article{chen2022wef,
  title={Why do we need large batchsizes in contrastive learning? a gradient-bias perspective},
  author={Chen, Changyou and Zhang, Jianyi and Xu, Yi and Chen, Liqun and Duan, Jiali and Chen, Yiran and Tran, Son and Zeng, Belinda and Chilimbi, Trishul},
  journal={Advances in Neural Information Processing Systems},
  volume={35},
  pages={33860--33875},
  year={2022}
}

@inproceedings{deng2009imagenet,
  title={Imagenet: A large-scale hierarchical image database},
  author={Deng, Jia and Dong, Wei and Socher, Richard and Li, Li-Jia and Li, Kai and Fei-Fei, Li},
  booktitle={2009 IEEE conference on computer vision and pattern recognition},
  pages={248--255},
  year={2009},
  organization={Ieee}
}

@article{cite9,
  author={Zhang, Xichen and Dadkhah, Sajjad and Weismann, Alexander Gerald and Kanaani, Mohammad Amin and Ghorbani, Ali A.},
  journal={IEEE Transactions on Computational Social Systems}, 
  title={Multimodal Fake News Analysis Based on Image–Text Similarity}, 
  year={2024},
  volume={11},
  number={1},
  pages={959-972},
  doi={10.1109/TCSS.2023.3244068}}

@inproceedings{wang2020cnn,
    title={CNN-Generated Images Are Surprisingly Easy to Spot... for Now.},
  author={Wang, Sheng-Yu and Wang, Oliver and Zhang, Richard and Owens, Andrew and Efros, Alexei A},
  booktitle={Proceedings of the IEEE/CVF conference on computer vision and pattern recognition},
  pages={8695--8704},
  year={2020}
}

@article{cite10,
  title={Temporal quality degradation in AI models},
  author={Vela, Daniel and Sharp, Andrew and Zhang, Richard and Nguyen, Trang and Hoang, An and Pianykh, Oleg S},
  journal={Scientific reports},
  volume={12},
  number={1},
  pages={11654},
  year={2022},
  publisher={Nature Publishing Group UK London},
  doi={10.1038/s41598-022-15245-z}
}

@inproceedings{cioni2024clip,
  title={Are CLIP features all you need for Universal Synthetic Image Origin Attribution?},
  author={Cioni, Dario and Tzelepis, Christos and Seidenari, Lorenzo and Patras, Ioannis},
  booktitle={European Conference on Computer Vision},
  pages={363--382},
  year={2024},
  organization={Springer}
}

@inproceedings{yang2023progressive,
  title={Progressive open space expansion for open-set model attribution},
  author={Yang, Tianyun and Wang, Danding and Tang, Fan and Zhao, Xinying and Cao, Juan and Tang, Sheng},
  booktitle={Proceedings of the IEEE/CVF Conference on Computer Vision and Pattern Recognition},
  pages={15856--15865},
  year={2023}
}

@inproceedings{yang2022deepfake,
  title={Deepfake network architecture attribution},
  author={Yang, Tianyun and Huang, Ziyao and Cao, Juan and Li, Lei and Li, Xirong},
  booktitle={Proceedings of the AAAI Conference on Artificial Intelligence},
  volume={36},
  pages={4662--4670},
  year={2022}
}

@article{dhamija2018reducing,
  title={Reducing network agnostophobia},
  author={Dhamija, Akshay Raj and G{\"u}nther, Manuel and Boult, Terrance},
  journal={Advances in Neural Information Processing Systems},
  volume={31},
  year={2018}
}

@inproceedings{vaze2021open,
  title={Open-Set Recognition: A Good Closed-Set Classifier is All You Need?},
  author={Vaze, S and Han, K and Vedaldi, A and Zisserman, A},
  booktitle={International Conference on Learning Representations (ICLR)},
  year={2022}
}

@inproceedings{bui2022repmix,
  title={Repmix: Representation mixing for robust attribution of synthesized images},
  author={Bui, Tu and Yu, Ning and Collomosse, John},
  booktitle={European Conference on Computer Vision},
  pages={146--163},
  year={2022},
  organization={Springer}
}

@InProceedings{cite12,
    author    = {Cozzolino, Davide and Poggi, Giovanni and Corvi, Riccardo and Nie{\ss}ner, Matthias and Verdoliva, Luisa},
    title     = {Raising the Bar of AI-generated Image Detection with CLIP},
    booktitle = {Proceedings of the IEEE/CVF Conference on Computer Vision and Pattern Recognition (CVPR) Workshops},
    month     = {June},
    year      = {2024},
    pages     = {4356-4366},
    doi ={10.1109/CVPRW63382.2024.00439}
}

@inproceedings{grad,
  title={Learning on gradients: Generalized artifacts representation for gan-generated images detection},
  author={Tan, Chuangchuang and Zhao, Yao and Wei, Shikui and Gu, Guanghua and Wei, Yunchao},
  booktitle={Proceedings of the IEEE/CVF Conference on Computer Vision and Pattern Recognition},
  pages={12105--12114},
  year={2023},
  doi={10.1109/CVPR52729.2023.01165c}
}

@article{zhao2024review,
  title={A review of convolutional neural networks in computer vision},
  author={Zhao, Xia and Wang, Limin and Zhang, Yufei and Han, Xuming and Deveci, Muhammet and Parmar, Milan},
  journal={Artificial Intelligence Review},
  volume={57},
  number={4},
  pages={99},
  year={2024},
  publisher={Springer}
}

@article{adv,
  title={Contrastive representation learning: A framework and review},
  author={Le-Khac, Phuc H and Healy, Graham and Smeaton, Alan F},
  journal={Ieee Access},
  volume={8},
  pages={193907--193934},
  year={2020},
  publisher={IEEE},
  doi={10.1109/ACCESS.2020.3031549}
}

@article{cruciani2020feature,
  title={Feature learning for human activity recognition using convolutional neural networks: A case study for inertial measurement unit and audio data},
  author={Cruciani, Federico and Vafeiadis, Anastasios and Nugent, Chris and Cleland, Ian and McCullagh, Paul and Votis, Konstantinos and Giakoumis, Dimitrios and Tzovaras, Dimitrios and Chen, Liming and Hamzaoui, Raouf},
  journal={CCF Transactions on Pervasive Computing and Interaction},
  volume={2},
  number={1},
  pages={18--32},
  year={2020},
  publisher={Springer}
}

@inproceedings{corvi2023intriguing,
  title={Intriguing properties of synthetic images: from generative adversarial networks to diffusion models},
  author={Corvi, Riccardo and Cozzolino, Davide and Poggi, Giovanni and Nagano, Koki and Verdoliva, Luisa},
  booktitle={Proceedings of the IEEE/CVF conference on computer vision and pattern recognition},
  pages={973--982},
  year={2023}
}

@inproceedings{xuan2019generalization,
  title={On the generalization of GAN image forensics},
  author={Xuan, Xinsheng and Peng, Bo and Wang, Wei and Dong, Jing},
  booktitle={Chinese conference on biometric recognition},
  pages={134--141},
  year={2019},
  organization={Springer}
}

@inproceedings{zhang2019detecting,
  title={Detecting and simulating artifacts in gan fake images},
  author={Zhang, Xu and Karaman, Svebor and Chang, Shih-Fu},
  booktitle={2019 IEEE international workshop on information forensics and security (WIFS)},
  pages={1--6},
  year={2019},
  organization={IEEE}
}

@article{tariang2024synthetic,
  title={Synthetic image verification in the era of generative artificial intelligence: What works and what isn’t there yet},
  author={Tariang, Diangarti and Corvi, Riccardo and Cozzolino, Davide and Poggi, Giovanni and Nagano, Koki and Verdoliva, Luisa},
  journal={IEEE Security \& Privacy},
  volume={22},
  number={3},
  pages={37--49},
  year={2024},
  publisher={IEEE}
}

@inproceedings{contr,
  title={Understanding the properties and limitations of contrastive learning for Out-of-Distribution detection},
  author={Keshtmand, Nawid and Santos-Rodriguez, Raul and Lawry, Jonathan},
  booktitle={International Conference on Pattern Recognition},
  pages={330--343},
  year={2022},
  organization={Springer},
  doi={10.1007/978-3-031-37660-3_23}
}

@article{cite13,
  title={Artifact feature purification for cross-domain detection of AI-generated images},
  author={Meng, Zheling and Peng, Bo and Dong, Jing and Tan, Tieniu and Cheng, Haonan},
  journal={Computer Vision and Image Understanding},
  volume={247},
  pages={104078},
  year={2024},
  publisher={Elsevier},
  doi ={10.1016/j.cviu.2024.104078}
}

@inproceedings{xu2025detecting,
  title={Detecting Origin Attribution for Text-to-Image Diffusion Models},
  author={Xu, Katherine and Zhang, Lingzhi and Shi, Jianbo},
  booktitle={2025 IEEE/CVF Winter Conference on Applications of Computer Vision (WACV)},
  pages={8775--8785},
  year={2025},
  organization={IEEE}
}

@article{cite14,
  title={Genimage: A million-scale benchmark for detecting ai-generated image},
  author={Zhu, Mingjian and Chen, Hanting and Yan, Qiangyu and Huang, Xudong and Lin, Guanyu and Li, Wei and Tu, Zhijun and Hu, Hailin and Hu, Jie and Wang, Yunhe},
  journal={Advances in Neural Information Processing Systems},
  volume={36},
  pages={77771--77782},
  year={2023},
  doi ={10.48550/arXiv.2306.08571}
}

@article{khosla2020supervised,
  title={Supervised contrastive learning},
  author={Khosla, Prannay and Teterwak, Piotr and Wang, Chen and Sarna, Aaron and Tian, Yonglong and Isola, Phillip and Maschinot, Aaron and Liu, Ce and Krishnan, Dilip},
  journal={Advances in neural information processing systems},
  volume={33},
  pages={18661--18673},
  year={2020}
}

@inproceedings{cite15,
  title={Dire for diffusion-generated image detection},
  author={Wang, Zhendong and Bao, Jianmin and Zhou, Wengang and Wang, Weilun and Hu, Hezhen and Chen, Hong and Li, Houqiang},
  booktitle={Proceedings of the IEEE/CVF International Conference on Computer Vision},
  pages={22445--22455},
  year={2023},
  doi ={10.1109/ICCV51070.2023.02051}
}

@inproceedings{clip,
  title={Learning transferable visual models from natural language supervision},
  author={Radford, Alec and Kim, Jong Wook and Hallacy, Chris and Ramesh, Aditya and Goh, Gabriel and Agarwal, Sandhini and Sastry, Girish and Askell, Amanda and Mishkin, Pamela and Clark, Jack and others},
  booktitle={International conference on machine learning},
  pages={8748--8763},
  year={2021},
  organization={PmLR},
  doi={10.48550/arXiv.2103.00020}
}

@inproceedings{npr,
  title={Rethinking the up-sampling operations in cnn-based generative network for generalizable deepfake detection},
  author={Tan, Chuangchuang and Zhao, Yao and Wei, Shikui and Gu, Guanghua and Liu, Ping and Wei, Yunchao},
  booktitle={Proceedings of the IEEE/CVF Conference on Computer Vision and Pattern Recognition},
  pages={28130--28139},
  year={2024},
  doi ={10.48550/arXiv.2312.10461}
}

@inproceedings{efficientnet,
  title={Efficientnet: Rethinking model scaling for convolutional neural networks},
  author={Tan, Mingxing and Le, Quoc},
  booktitle={International conference on machine learning},
  pages={6105--6114},
  year={2019},
  organization={PMLR},
 doi = {10.48550/arXiv.1905.11946}
}

@article{cite16,
  author={Xu, Juncong and Yang, Yang and Fang, Han and Liu, Honggu and Zhang, Weiming},
  journal={IEEE Signal Processing Letters}, 
  title={FAMSeC: A Few-Shot-Sample-Based General AI-Generated Image Detection Method}, 
  year={2025},
  volume={32},
  number={},
  pages={226-230},
  doi={10.1109/LSP.2024.3511421}
}

@article{cite19,
  title={Cifake: Image classification and explainable identification of ai-generated synthetic images},
  author={Bird, Jordan J and Lotfi, Ahmad},
  journal={IEEE Access},
  volume={12},
  pages={15642--15650},
  year={2024},
  publisher={IEEE},
  doi={10.1109/ACCESS.2024.3356122}
}

@inproceedings{cite20,
  title={Detecting generated images by real images},
  author={Liu, Bo and Yang, Fan and Bi, Xiuli and Xiao, Bin and Li, Weisheng and Gao, Xinbo},
  booktitle={European Conference on Computer Vision},
  pages={95--110},
  year={2022},
  organization={Springer},
  doi={10.1007/978-3-031-19781-9_6}
}

@inproceedings{cite21,
  title={Deep image fingerprint: Towards low budget synthetic image detection and model lineage analysis},
  author={Sinitsa, Sergey and Fried, Ohad},
  booktitle={Proceedings of the IEEE/CVF Winter Conference on Applications of Computer Vision},
  pages={4067--4076},
  year={2024},
  doi={10.1109/WACV57701.2024.00402}
}

@article{cite28,
  title={Face forgery detection via multi-feature fusion and local enhancement},
  author={Zhang, Dengyong and Chen, Jiahao and Liao, Xin and Li, Feng and Chen, Jiaxin and Yang, Gaobo},
  journal={IEEE Transactions on Circuits and Systems for Video Technology},
  year={2024},
  publisher={IEEE},
  doi={10.1109/TCSVT.2024.3390945}
}

@inproceedings{cite29,
  title={De-fake: Detection and attribution of fake images generated by text-to-image generation models},
  author={Sha, Zeyang and Li, Zheng and Yu, Ning and Zhang, Yang},
  booktitle={Proceedings of the 2023 ACM SIGSAC conference on computer and communications security},
  pages={3418--3432},
  year={2023},
  doi={10.1145/3576915.3616588}
}

@InProceedings{cite30,
    author    = {Epstein, David C. and Jain, Ishan and Wang, Oliver and Zhang, Richard},
    title     = {Online Detection of AI-Generated Images},
    booktitle = {Proceedings of the IEEE/CVF International Conference on Computer Vision (ICCV) Workshops},
    month     = {October},
    year      = {2023},
    pages     = {382-392},
   doi ={10.1109/ICCVW60793.2023.00045}
}

@inproceedings{cite31,
  title={Attributing fake images to gans: Learning and analyzing gan fingerprints},
  author={Yu, Ning and Davis, Larry S and Fritz, Mario},
  booktitle={Proceedings of the IEEE/CVF international conference on computer vision},
  pages={7556--7566},
  year={2019},
  doi={10.1109/ICCV.2019.00765}
}

@inproceedings{cite32,
  title={Did you use my gan to generate fake? post-hoc attribution of gan generated images via latent recovery},
  author={Hirofumi, Syou and Fukuchi, Kazuto and Akimoto, Yohei and Sakuma, Jun},
  booktitle={2022 International Joint Conference on Neural Networks (IJCNN)},
  pages={1--8},
  year={2022},
  organization={IEEE},
  doi={10.1109/IJCNN55064.2022.9892704}
}

@incollection{cite33,
title = {Chapter 6 - A review of techniques to detect the GAN-generated fake images},
editor = {Arun Solanki and Anand Nayyar and Mohd Naved},
booktitle = {Generative Adversarial Networks for Image-to-Image Translation},
publisher = {Academic Press},
pages = {125-159},
year = {2021},
isbn = {978-0-12-823519-5},
doi = {https://doi.org/10.1016/B978-0-12-823519-5.00004-X},
author = {Tanvi Arora and Rituraj Soni},
}

@article{cite34,
  title={Detecting image attribution for text-to-image diffusion models in rgb and beyond},
  author={Xu, Katherine and Zhang, Lingzhi and Shi, Jianbo},
  journal={arXiv preprint arXiv:2403.19653},
  year={2024},
  doi={10.48550/arXiv.2403.19653}
}

@article{cite62,
  title={Sounds of Science: Copyright Infringement in AI Music Generator Outputs},
  author={Sunray, Eric},
  journal={Catholic University Journal of Law and Technology},
  volume={29},
  number={2},
  pages={185--218},
  year={2021}
}

@article{bigGan,
  title={Large scale GAN training for high fidelity natural image synthesis},
  author={Brock, Andrew and Donahue, Jeff and Simonyan, Karen},
  journal={arXiv preprint arXiv:1809.11096},
  year={2018},
  doi={10.48550/arXiv.1809.11096}
}

\end{document}